\pdfoutput=1
\documentclass[11pt]{article}
\usepackage[a4paper,margin=2.5cm]{geometry}
\usepackage[T1]{fontenc}
\usepackage{lmodern}
\usepackage{microtype}
\usepackage{booktabs}
\usepackage{array}
\usepackage{amsmath}
\usepackage{graphicx}
\usepackage{xcolor}
\usepackage{tikz}
\usetikzlibrary{arrows.meta}
\usepackage{pgfplots}
\pgfplotsset{compat=1.18}
\definecolor{figblue}{HTML}{2A78D6}
\definecolor{figorange}{HTML}{EB6834}
\definecolor{figaqua}{HTML}{1BAF7A}
\usepackage{xurl}
\usepackage[hidelinks]{hyperref}
\usepackage{adjustbox}
\usepackage{float}
\usepackage{placeins} % floats stay inside their own section
\usepackage{caption}
\usepackage{natbib}
\setcitestyle{numbers,square}

\title{\textbf{Beyond Word Error Rate: A Failure Taxonomy of Streaming Dictation ASR}}
\author{%
  \begin{tabular}{@{}c@{\hspace{3em}}c@{}}
  Harras Mansoor & Muhammad Wisal \\
  {\small\texttt{harrasmansoor3@gmail.com}} & {\small\texttt{muhammadwisalkhanmv@gmail.com}} \\[1.2ex]
  Hassan Bhatti & Masood Shakir \\
  {\small\texttt{hassan.w.bhatti@gmail.com}} & {\small\texttt{shakirmalik2724@gmail.com}} \\[1.6ex]
  \multicolumn{2}{c}{heylemon.ai}
  \end{tabular}}
\date{}

\begin{document}
\maketitle

\begin{abstract}
Speech recognition (ASR) systems are usually compared by word error rate (WER) on recorded files, after punctuation, capital letters and number format are removed. In dictation, the audio is streamed live and the text goes into the user's message as written, so what matters is which mistakes the user has to fix. The test set, streamed to eleven ASR systems, comprises 106 scripted English dictations from one speaker (read in a quiet room, with a fan on, and whispered), 20 clips in each of eleven other language and dialect sets, and 400 recordings that join clips in two languages. Every error was sorted by a fixed rule into four levels: recognition, output formatting, language handling and, for the self-hosted models, audio dropped before the model. Systems with similar WER failed in different ways. Zoom Scribe and OpenAI's GPT-Live-Transcribe have almost the same WER on quiet English (7.4\% and 7.3\%), but 34.7\% and 8.8\% on whispered English. Zoom and both AssemblyAI models left out 26--28\% of whispered words, and still 21--26\% with their end-of-turn detection relaxed or off; seven other systems left out at most 3\%. Qwen3-ASR's 32\% came from the speech detection in our own deployment (6\% when relaxed). Across 107 EARS speakers, the same four systems lost 7--31\% of whispered words, while six others lost under 2\%. With only English selected, Gemini, Deepgram and our self-hosted Whisper lost the non-English part of a two-language recording in 86--99\% of recordings, against 33--46\% with both languages selected, while GPT-Live-Transcribe, OpenAI's realtime Whisper and Meta lost it in at most 20\%. Running Whisper behind a Silero voice activity detector that drops audio judged to be silence raised whispered WER from 10.1\% to 51.4\%. Most English results come from one read speaker. The code, the recordings, both references and every transcript are released.
\end{abstract}

\section{Introduction}
This paper studies one use of ASR systems, referred to here as \emph{dictation}: the user speaks a message, an email, a note or an instruction to an AI assistant, and the text is placed into another app. The transcript is the final product. Every error stays in the message unless the user finds it and corrects it, and correcting recognition errors is a large part of the cost of entering text by voice \citep{vertanen2010parakeet,ruan2017speech}.

WER counts errors but does not say what kind they are. On a dictation of eleven sentences of similar length, a WER of 9\% could mean one wrong word in every eleven, or one whole sentence missing from otherwise correct text. A single average can also hide large differences between groups of speakers \citep{koenecke2020racial} and between dialects \citep{talafha2024casablanca}.

Public benchmarks run complete audio files through each system and report WER on normalised text \citep{radford2023whisper,srivastav2026openasr}. Dictation differs in four ways, each tested here:
\begin{itemize}\itemsep0pt
  \item The audio is streamed, so it is streamed at the speed of speech and the time to the final text after the audio ends is measured.
  \item The text is used as written, so numbers are scored as written and errors in punctuation and capital letters are also reported.
  \item People dictate in different conditions, so English is recorded in a quiet room, with a fan on and in a whisper.
  \item People dictate in other languages and mix languages, so 11 language and dialect sets covering 7 languages are tested, along with recordings that switch from one language to another.
\end{itemize}

Eleven systems are tested: two self-hosted open models and nine services run by their providers. Every transcript is lined up with the correct text word by word, and each error is sorted by a fixed rule into one of four groups, called \emph{levels}: recognition, output formatting, language handling, and the audio path before the model (Section~\ref{sec:taxonomy}).

The main finding is that systems with similar WER fail in different ways, and that the differences are large. Zoom and GPT-Live have almost the same quiet-English WER (7.4\% and 7.3\%), but on whispered English Zoom leaves out 28\% of words and GPT-Live 1\%. With only English selected, some systems remove the other language of a two-language recording almost every time and others keep it. And the app around the model matters: running Whisper behind a voice activity detector that drops audio it judges to be silence raised its whispered WER from 10.1\% to 51.4\%.

The paper makes four contributions:
\begin{itemize}\itemsep0pt
  \item \textbf{A failure taxonomy for streaming dictation} that groups error types by the level they show up at, with a stated rule for detecting each type (Section~\ref{sec:taxonomy}).
  \item \textbf{A per-level comparison of eleven systems} on in-house data and on two external corpora with many speakers, including two behaviours known for open models \citep{peng2023prompting,rahman2026scriptcollapse,azeemi2026southasian} that, to our knowledge, have not been measured for commercial streaming systems: a language setting that removes the other language for some systems and only guides the language for others, and output in the wrong writing system (Sections~\ref{sec:overall}--\ref{sec:cs}).
  \item \textbf{Two references for the same speech}: the script the speaker meant (the \emph{script reference}) and a record of what they said (the \emph{spoken reference}), which separate a system that hears every word from one that writes the intended message (Section~\ref{sec:cleanup}).
  \item \textbf{A case study of the level before the model}: nine ways of running Whisper that differ in hosting, in how the audio is cut, in which voice activity detector (VAD) is used, and in whether the VAD drops audio before the model (Section~\ref{sec:whisperguide}).
\end{itemize}
The test client, the analysis code, the English recordings, both references and every system's transcripts are released.

\FloatBarrier
\section{Related Work}\label{sec:related}
\paragraph{ASR benchmarks and error analysis.} Public leaderboards test open models and commercial services on complete audio files from public datasets, and clean up the text first so that differences in formatting do not count as errors \citep{srivastav2026openasr,radford2023whisper}. Independent tests of commercial services stream audio at the speed of speech and report speed alongside accuracy \citep{aa}. WER depends on how the text is normalised and on the test data \citep{szymanski2020wer}, and WER on one test set is a poor guide to WER on others \citep{likhomanenko2021rethinking}. Other measures look past WER: match error rate and word information lost \citep{morris2004wer}, semantic distance between reference and transcript \citep{kim2021semdist}, scores that split hallucinations by kind \citep{koudounas2026shallow}, and rates of runs of inserted or deleted words per hour \citep{ramirez2024anatomy}. Error analysis has shown which words are hard to recognise, such as rare words and words near disfluencies \citep{goldwater2010words}, and work on detecting and correcting ASR errors is reviewed by \citet{errattahi2018errors}. Error categories close to ours have been applied to Whisper on South Asian languages, where output in the wrong writing system caused most failures in three of four languages \citep{azeemi2026southasian}. Wrong-script output has also been measured across many model--language pairs \citep{rahman2026scriptcollapse} and removed from WER by mapping both texts to one script \citep{pattnayak2026snwer}. These studies test open models. Some benchmarks score text closer to how it is written: ESB keeps casing and punctuation \citep{gandhi2022esb}, LibriSpeech-PC scores punctuation and capital letters \citep{meister2023librispeechpc}, and Earnings-21 reports accuracy on named entities \citep{delrio2021earnings21}. Most of these report how many errors a system makes on complete files. This work reports what kinds of errors commercial systems make in a streaming dictation setting, scoring the text as the user receives it, without converting numbers to a common form.

\paragraph{Dictation and text entry.} In a study of short messages on phones, speech entry was about three times faster than typing, with slightly more errors left in the final text \citep{ruan2017speech}. The delay a user feels depends on how fast text is emitted and on endpointing, the decision that the speaker has stopped \citep{shangguan2021latency}, and partial text can change before it is final \citep{shangguan2020stability}.

\paragraph{Many languages and dialects.} FLEURS \citep{conneau2023fleurs} and VoxPopuli \citep{wang2021voxpopuli} provide read speech and parliament speeches in many languages. Arabic dialects remain much harder for current systems than Modern Standard Arabic \citep{talafha2024casablanca,alharbi2024sada}. Arabic dialects also have no standard spelling, which has led to spelling conventions \citep{habash2012coda} and to scoring against several correct texts \citep{ali2015multireference}. Urdu ASR has been benchmarked on read and conversational speech \citep{arif2025werwestand}. Urdu is also often written in Latin letters (Roman Urdu), and converting code-mixed Roman Urdu back to Urdu script is a task of its own \citep{wisal2022cmrutu}.

\paragraph{Switching languages.} Code-switching, moving between languages between or within sentences \citep{poplack1980sometimes}, has long been hard for speech and language technology \citep{winata2023decades}. Existing corpora, such as SEAME \citep{lyu2010seame} and ASCEND \citep{lovenia2022ascend} for Mandarin and English, contain natural switches within a sentence, for a few language pairs. For Whisper, the language given in the prompt changes code-switched output, and a single language can lead it to write a translation instead of a transcript \citep{peng2023prompting}. WER over a whole recording can hide errors on the switched words, which has led to scoring only those words \citep{ugan2025pier}. The recordings used here instead switch between sentences, for many language pairs. These recordings are less natural, but they reduce the chance that a language was lost because it is hard on its own.

\paragraph{Whispered speech.} Whispered speech has no voicing and is much harder for recognisers trained on normal speech \citep{lim2011whisper}; work on whispered ASR adapts models to it, for example with synthetic pseudo-whisper training data \citep{ghaffarzadegan2016whisper}. Corpora with whispered readings include wTIMIT \citep{lim2011whisper}, CHAINS \citep{cummins2006chains}, AISHELL6-Whisper for Mandarin \citep{li2026aishell6whisper} and EARS \citep{richter2024ears}, which records the same text read normally and whispered by 107 speakers; EARS is used here to compare the systems across speakers.

\FloatBarrier
\section{Failure Taxonomy}\label{sec:taxonomy}
\begin{table}[!htbp]
\centering\small
\begin{adjustbox}{max width=\textwidth}
\begin{tabular}{@{}>{\raggedright\arraybackslash}p{2.4cm}>{\raggedright\arraybackslash}p{3.6cm}>{\raggedright\arraybackslash}p{8.2cm}r@{}}
\toprule
Level & Error type & Rule & Section \\
\midrule
Recognition & Wrong, missing, extra word (substitution, deletion, insertion) & a word replaced by another, left out, or added, from the word-by-word comparison & \ref{sec:mix} \\
& Near-miss & a wrong word of 4 or more letters that differs from the correct word by 1 or 2 letters & \ref{sec:mix} \\
& Name error & a word of a name in the correct text is not in the transcript & \ref{sec:names} \\
& Missing stretch & 5 or more words in a row are missing (a run of deletions, as in the omission rate of \citealp{ramirez2024anatomy}), noted by position (first, middle or last third) & \ref{sec:whisper} \\
& Cut-off ending & the last 3 or more words are missing (truncation) & \ref{sec:whisper} \\
& Failed / lost clip & the clip's WER is at least 50\% / at least 90\% & \ref{sec:whisper}, \ref{sec:dialects} \\
& Hard clip & the clip's WER is above 30\% & \ref{sec:dialects} \\
& Invented text (hallucination) & text written for a clip with no speech, or a short stock phrase written in place of a whole clip's speech & \ref{sec:whisperguide} \\
\addlinespace
Output formatting & Number format (ITN) & a number in digits in the correct text comes out in words, or the other way round & \ref{sec:numbers} \\
& Formatting & a wrong capital letter or punctuation mark (formatted error rate, Table~\ref{tab:overall}) & \ref{sec:overall} \\
& Clean-up (disfluency removal) & fillers, repeats and self-corrections removed: the difference between WER against the spoken reference and against the script reference & \ref{sec:cleanup} \\
& Wrong writing system & most of the letters in the text belong to a different writing system from the chosen language's & \ref{sec:script} \\
\addlinespace
Language handling & Missing part & less than 20\% of the words of one language's part of a two-language recording are heard & \ref{sec:cs} \\
& Romanised part (transliteration) & a missing part is replaced by Latin-letter text that is not English & \ref{sec:cs} \\
& Translated part & a missing part is replaced by English (at least 4 words, at least 15\% of them common English function words such as ``the'' or ``of'') & \ref{sec:cs} \\
\addlinespace
Before the model & Dropped by the app's VAD (VAD gating) & audio the app never sent to the system & \ref{sec:whisperguide} \\
\bottomrule
\end{tabular}
\end{adjustbox}
\caption{The failure taxonomy: error types grouped by the level they show up at, with the field's usual term in brackets, the rule used to detect each, and the section that measures it. The 50\% bar for a failed clip is where half of the message would have to be retyped; the 30\% bar for a hard clip counts a clip that needs several corrections; the 20\% bar keeps a part with a few misheard words from counting as missing.}
\label{tab:taxonomy}
\end{table}
Error types are grouped into four levels, by where in the pipeline the error shows up (Table~\ref{tab:taxonomy}): \textbf{recognition} (words missed, misheard or invented); \textbf{output formatting} (words recognised but written in a form the user did not intend, such as a number in words instead of digits, or text in the wrong writing system; \emph{inverse text normalisation}, ITN, is the step that turns ``seven'' into ``7'' \citep{pusateri2017itn}); \textbf{language handling} (one language of a two-language message dropped, written in Latin letters, or translated); and \textbf{before the model} (audio the app never sends to the system).
The levels describe what is observed in the output, not a proven cause. For the hosted services, a missing word could have been lost in the service's own speech detection rather than in its model, and a language could be lost because of when the service finalises text rather than how it handles languages (Section~\ref{sec:cs}). The level before the model can be separated from recognition only for the self-hosted models, Whisper and Qwen, whose audio path is under our control (Sections~\ref{sec:whisper} and~\ref{sec:whisperguide}). Every result therefore describes a system together with our integration of it (Appendix~\ref{app:systems}).

Each type has a stated rule that a program applies to the word-by-word comparison; the rules that rely on earlier judgements are listed in Section~\ref{sec:limits}.

\FloatBarrier
\section{Experimental Setup}\label{sec:setup}
\paragraph{Systems.} Eleven systems that transcribe in real time are tested. Two are self-hosted open models: Qwen3-ASR 1.7B \citep{qwen3asr} (``Qwen'') and Whisper Large V3 \citep{radford2023whisper} (``Whisper SH'', self-hosted). Nine are run by their providers: Meta Muse Voice Transcribe (``Meta''); Gemini 3.5 Transcribe (Live API; ``Gemini''); Deepgram Nova-3; AssemblyAI Universal-3.6 Pro and Universal-3.5 Pro (``AAI 3.6'' and ``AAI 3.5''); Zoom Scribe (Live API; ``Zoom''); ElevenLabs Scribe v2 Realtime (``ElevenLabs''); OpenAI GPT-Live-Transcribe (Realtime API; ``GPT-Live''); and OpenAI GPT-Realtime-Whisper, OpenAI's hosted streaming Whisper (``RT Whisper''). Section~\ref{sec:whisperguide} compares nine ways of running Whisper; the main comparison uses two of them, Whisper SH and RT Whisper. All runs took place between 1 October and 5 November 2026 (Appendix~\ref{app:systems}).

Each system is used with its default settings, with a few changes given in Appendix~\ref{app:systems}. Each system is told the language a user would choose, as a language code without a region unless a system needs one; if a system has no option for that language, none is sent and the system detects the language itself. Sending each dialect's own regional code, where a provider documents one, changed almost nothing (Appendix~\ref{app:systems}).

\paragraph{Disfluency-removal settings.} Gemini and ElevenLabs write every word as spoken by default, including fillers, repeats and false starts, and each offers a setting that removes them: Gemini's Smart mode, which also adds punctuation, capital letters and structured formatting \citep{geminitranscribedocs}, and ElevenLabs' \texttt{no\_verbatim} option \citep{elevenlabsdocs}. These are reported as two extra rows, ``Gemini Smart'' and ``ElevenLabs NV'' (no verbatim), so the result tables have thirteen rows. Counts of errors that all systems share cover the eleven systems and leave these two rows out.

\paragraph{In-house English data.} The in-house set consists of 106 English messages, written for the study to resemble chat, email, notes and prompts to AI assistants, and read aloud by one speaker, a member of the research team. Each message was recorded once, in one condition: 73 in a quiet room, 17 with a fan running nearby and 16 in a whisper. The three conditions use different messages, so systems are compared within a condition, not the conditions with each other; the fan-on set is a second set of read messages, not a test of noise. The third condition is called \emph{whispered speech}; the model is always named by its setup (for example Whisper SH), to keep the two apart. The reader deliberately moved between US, South Asian, British and Irish English, often within a message (Appendix~\ref{app:accent}).

\paragraph{Script and spoken references.} Each English message has two correct texts. The \emph{script reference} is the text the reader was given: what they meant to send. The \emph{spoken reference} is what they actually said, made by correcting the script against the audio. The two differ in 85 of the 106 messages, by 453 of 10{,}776 words (4.2\%): 242 added, 88 left out and 123 changed, including repeated words and spoken self-corrections. To make the spoken reference, a program proposed an edit wherever several systems heard the same departure from the script, and one listener accepted or rejected each proposal and corrected the rest by hand (Appendix~\ref{app:data}). Unless stated otherwise, English results use the script reference, because it is what a dictation user wants on the screen; Section~\ref{sec:cleanup} compares the two.

\paragraph{Multilingual, code-switched and no-speech data.} Twenty short clips were taken from each of 11 language and dialect sets in public datasets (Table~\ref{tab:data}). From these, 400 recordings were built that join two clips in different languages (Section~\ref{sec:cs}). Ten clips with no speech are also included (silence and background noise) to check whether a system writes text when nobody is speaking. The single-language clips add up to 99 minutes of speech.

\paragraph{External corpora.} Two public corpora with many speakers repeat the whispering and noise tests: EARS and VOiCES (Sections~\ref{sec:whisper} and~\ref{sec:voices}, Appendix~\ref{app:ext}). They are kept apart from the in-house data and scored differently.

\begin{table}[!htbp]
\centering\small
\begin{adjustbox}{max width=\textwidth}
\begin{tabular}{@{}llr@{}}
\toprule
Set & Source & Clips \\
\midrule
English, quiet & in-house & 73 \\
English, fan on & in-house & 17 \\
English, whispered & in-house & 16 \\
\midrule
Arabic (Modern Standard, MSA; Egyptian speakers) & FLEURS \citep{conneau2023fleurs} & 20 \\
Arabic (Saudi: 12 Najdi and 8 Hijazi clips) & SADA \citep{alharbi2024sada} & 20 \\
Arabic (Emirati) & Casablanca \citep{talafha2024casablanca} & 20 \\
German, French, Hindi, Urdu & FLEURS & 4 $\times$ 20 \\
Mandarin Chinese & FLEURS & 20 \\
Spanish (Latin American) & FLEURS & 20 \\
Spanish (Spain) & VoxPopuli \citep{wang2021voxpopuli} & 20 \\
Spanish (Puerto Rico) & OpenSLR \citep{guevara2020crowdsourcing} & 20 \\
\midrule
No speech & in-house & 10 \\
Two-language recordings & built from the clips above & 400 \\
\midrule
External: whispered and normal & EARS \citep{richter2024ears} & 2 $\times$ 428 \\
External: clean, far, far + babble, far + TV & VOiCES \citep{richey2018voices} & 4 $\times$ 100 \\
\bottomrule
\end{tabular}
\end{adjustbox}
\caption{The test data: number of clips or recordings in each set.}
\label{tab:data}
\end{table}

\paragraph{Streaming protocol.} Each clip is streamed at the speed of speech in small chunks, the way a microphone delivers it, with a new session for every clip and all systems called from the same client machine (Appendix~\ref{app:systems}). A system \emph{finalises} text when it commits to it and stops revising it; only finalised text is scored.

A voice activity detector (VAD) is a small model that decides whether audio contains speech. Except where Section~\ref{sec:whisperguide} says otherwise, no VAD drops any audio: every system receives every sample. Whisper SH uses Silero VAD \citep{silero} only to decide where to cut the audio into segments. Section~\ref{sec:whisperguide} measures what happens when a VAD also drops audio, using a second Whisper deployment of ours with a VAD in front of it on the client (``gated Whisper'').

\paragraph{Scoring.} Before a transcript is compared with the correct text, capital letters and punctuation are ignored, and variant letter forms in Arabic, Urdu and Chinese are treated as equal (Appendix~\ref{app:data}). Numbers are not converted: the text is scored as written, so ``seven'' in place of ``7'' counts as an error.

Each transcript is lined up with the correct text (the reference) using the fewest possible edits (a Levenshtein alignment). This gives three kinds of word error: a \emph{substitution} (a wrong word), a \emph{deletion} (a missing word) and an \emph{insertion} (an extra word). WER is the total number of errors in a set divided by the total number of words in its correct text. It can exceed 100\%, because extra words also count as errors. Mandarin, which is not written with spaces, is scored by counting each Chinese character as one unit (Appendix~\ref{app:data}). The \emph{character error rate} (CER) is the same measure counted over characters. A \emph{formatted} error rate is also reported, a case- and punctuation-sensitive WER as in ESB \citep{gandhi2022esb}: words keep their capital letters and each punctuation mark counts as a separate token, so a wrong capital or a missing comma each counts as one error.

\paragraph{Uncertainty.} English WER is given with a 95\% confidence interval from a percentile bootstrap over clips with 2{,}000 resamples \citep{bisani2004bootstrap} (4{,}000 for the significance tests). These intervals describe variation across messages, not across speakers. When two systems, or one system under two conditions, are compared, the same clips are resampled for both (a paired comparison). A difference is \emph{clear} when the 95\% interval of the paired difference excludes zero and it still holds after a Holm correction for the number of comparisons in its group (for example, the 13 name gaps form one group; Appendix~\ref{app:data}); otherwise it is \emph{within chance}. The language sets have only 20 clips each, so differences of a few points there are within chance. Differences are given in percentage points (``points'').

\FloatBarrier
\section{Overall Results}\label{sec:overall}
\paragraph{Similar WER on quiet English, different WER elsewhere.} Table~\ref{tab:overall} gives each system's WER and formatted error rate for each English condition. On quiet English, ten of the thirteen rows are between 3.9\% and 9.5\%, and the order of these systems tells little about the other conditions. Zoom (7.4\%) and GPT-Live (7.3\%) are the clearest case: on whispered English, Zoom's WER is 34.7\% and GPT-Live's 8.8\% (paired difference 25.8 points, 95\% CI 10.3 to 51.7), almost all of it missing words. AAI 3.6 (8.2\%) and RT Whisper (8.2\%) are another: 30.5\% and 8.9\% whispered. Gemini Smart has the lowest quiet-English WER (3.9\%), clearly below the next row, ElevenLabs NV; the lowest with default settings is Gemini, at 6.9\%. When capital letters and punctuation are also scored, error rates on quiet English rise to 8.8--22.8\%; Gemini Smart (8.8\%), ElevenLabs NV (9.0\%) and Gemini (9.1\%) are lowest.

\begin{table}[!htbp]
\centering\small
\begin{adjustbox}{max width=\textwidth}
\begin{tabular}{@{}lrrrrrr@{}}
\toprule
& \multicolumn{3}{c}{WER} & \multicolumn{3}{c}{Formatted} \\
\cmidrule(lr){2-4}\cmidrule(l){5-7}
System & Quiet & Fan on & Whispered & Quiet & Fan on & Whispered \\
\midrule
Whisper SH & 8.3 {\scriptsize [6.8, 9.8]} & 6.5 {\scriptsize [4.1, 10.0]} & 10.1 {\scriptsize [7.1, 13.0]} & 16.0 & 14.3 & 20.8 \\
Qwen & 13.6 {\scriptsize [10.9, 17.0]} & 7.0 {\scriptsize [5.3, 9.1]} & 46.6 {\scriptsize [32.9, 67.5]} & 21.6 & 15.5 & 56.5 \\
Meta & 9.5 {\scriptsize [8.3, 10.6]} & 7.4 {\scriptsize [5.3, 9.6]} & 11.3 {\scriptsize [9.0, 14.1]} & 18.3 & 14.9 & 22.4 \\
Gemini & 6.9 {\scriptsize [5.8, 8.0]} & 2.5 {\scriptsize [1.8, 3.3]} & 5.0 {\scriptsize [2.8, 6.7]} & 9.1 & 6.4 & 8.6 \\
Gemini Smart & \textbf{3.9} {\scriptsize [3.2, 4.7]} & 3.2 {\scriptsize [2.1, 4.6]} & 4.2 {\scriptsize [2.9, 5.7]} & 8.8 & 10.0 & 10.9 \\
Deepgram & 13.6 {\scriptsize [12.0, 15.1]} & 11.4 {\scriptsize [8.8, 14.0]} & 19.8 {\scriptsize [15.7, 23.6]} & 22.8 & 20.0 & 32.1 \\
AAI 3.6 & 8.2 {\scriptsize [7.2, 9.1]} & 4.1 {\scriptsize [2.7, 5.7]} & 30.5 {\scriptsize [14.6, 55.8]} & 13.1 & 8.8 & 35.1 \\
AAI 3.5 & 9.0 {\scriptsize [7.9, 10.1]} & 4.3 {\scriptsize [2.8, 6.2]} & 30.9 {\scriptsize [15.2, 55.9]} & 14.9 & 9.4 & 35.9 \\
Zoom & 7.4 {\scriptsize [6.5, 8.3]} & 3.6 {\scriptsize [2.2, 5.2]} & 34.7 {\scriptsize [19.2, 60.2]} & 13.7 & 10.2 & 41.6 \\
ElevenLabs & 12.8 {\scriptsize [10.9, 14.9]} & 5.8 {\scriptsize [3.7, 8.2]} & 9.1 {\scriptsize [5.4, 12.9]} & 21.9 & 11.6 & 16.0 \\
ElevenLabs NV & 5.1 {\scriptsize [4.3, 5.9]} & 4.5 {\scriptsize [3.1, 6.0]} & 7.1 {\scriptsize [3.6, 10.7]} & 9.0 & 8.9 & 12.4 \\
GPT-Live & 7.3 {\scriptsize [6.3, 8.3]} & 3.6 {\scriptsize [2.4, 4.8]} & 8.8 {\scriptsize [7.0, 10.4]} & 11.2 & 8.1 & 14.2 \\
RT Whisper & 8.2 {\scriptsize [7.0, 9.5]} & 4.6 {\scriptsize [3.2, 6.4]} & 8.9 {\scriptsize [7.2, 11.0]} & 11.9 & 8.1 & 13.8 \\
\bottomrule
\end{tabular}
\end{adjustbox}
\caption{English error rates (\%), lower is better. WER is shown with its 95\% confidence interval. Bold: clearly below the next-lowest system (paired bootstrap); this holds only for quiet-English WER. Formatted: WER that also counts mistakes in capital letters and punctuation.}
\label{tab:overall}
\end{table}

\paragraph{Other languages.} For European languages and Mandarin the systems are close, and most stay below 10\%. Hindi ranges from 2.6\% to 34.6\%, Urdu is above 85\% for several systems (Section~\ref{sec:script}), and the Saudi and Emirati Arabic sets are hard for every system (Section~\ref{sec:dialects}). No system is clearly best on any language set. Appendix~\ref{app:extra} gives WER for each set (Table~\ref{tab:overalllang}).

\paragraph{Latency.} Measured from our test client, typical (median) waits from the end of the audio to the final text range from 313 to 760\,ms. Meta, Deepgram, Whisper SH, Qwen and both ElevenLabs settings wait under 400\,ms; GPT-Live and RT Whisper about 750\,ms. Appendix~\ref{app:speed} gives the full timings, which should be treated as indicative.

\FloatBarrier
\section{Recognition Errors}\label{sec:recognition}
\subsection{Substitutions, Deletions and Insertions}\label{sec:mix}
WER adds three kinds of error together: wrong words (substitutions), missing words (deletions) and extra words (insertions). Table~\ref{tab:fmix} splits each system's WER into these three parts for quiet English, English with the fan on, and the other languages.

\begin{table}[!htbp]
\centering\small
\begin{adjustbox}{max width=\textwidth}
\begin{tabular}{@{}lrrrrrrrrr@{}}
\toprule
& \multicolumn{3}{c}{Quiet English} & \multicolumn{3}{c}{English, fan on} & \multicolumn{3}{c}{Other languages} \\
\cmidrule(lr){2-4}\cmidrule(lr){5-7}\cmidrule(l){8-10}
System & Wrong & Missing & Extra & Wrong & Missing & Extra & Wrong & Missing & Extra \\
\midrule
Whisper SH & 3.2 & 2.2 & 2.9 & 4.2 & 0.8 & 1.5 & 12.1 & 3.7 & 0.9 \\
Qwen & 5.0 & 2.7 & 6.0 & 4.6 & 0.4 & 2.0 & 11.7 & 2.5 & 0.9 \\
Meta & 3.5 & 1.1 & 4.9 & 4.6 & 0.6 & 2.2 & 9.4 & 2.6 & 1.0 \\
Gemini & 2.1 & 1.1 & 3.7 & 1.6 & 0.5 & 0.5 & 7.8 & 2.0 & 0.9 \\
Gemini Smart & 1.9 & 1.3 & 0.7 & 1.8 & 1.1 & 0.3 & 7.6 & 3.2 & 0.6 \\
Deepgram & 5.8 & 2.5 & 5.3 & 7.0 & 1.8 & 2.6 & 9.2 & 5.7 & 0.9 \\
AAI 3.6 & 2.8 & 1.3 & 4.0 & 2.5 & 0.6 & 1.0 & 9.4 & 1.9 & 0.6 \\
AAI 3.5 & 2.9 & 1.4 & 4.7 & 2.5 & 0.8 & 1.1 & 12.1 & 2.2 & 0.7 \\
Zoom & 2.6 & 1.4 & 3.4 & 2.5 & 0.5 & 0.7 & 10.3 & 2.8 & 1.0 \\
ElevenLabs & 2.8 & 1.9 & 8.1 & 3.5 & 0.5 & 1.9 & 11.1 & 1.9 & 1.2 \\
ElevenLabs NV & 2.5 & 1.2 & 1.4 & 3.0 & 0.5 & 1.0 & 13.1 & 4.0 & 1.2 \\
GPT-Live & 2.9 & 1.2 & 3.2 & 2.5 & 0.5 & 0.6 & 8.6 & 3.6 & 0.6 \\
RT Whisper & 2.9 & 1.2 & 4.1 & 3.0 & 0.8 & 0.8 & 9.1 & 3.2 & 0.5 \\
\bottomrule
\end{tabular}
\end{adjustbox}
\caption{Wrong, missing and extra words (substitutions, deletions, insertions) against the script reference, each as a share (\%) of all words in the correct text. For each English condition, the three add up to the WER, give or take 0.1 for rounding. Other languages: the 10 non-English sets other than Urdu, pooled by words, so the sum differs slightly from the per-set average in Table~\ref{tab:overalllang} (Urdu errors are mostly the wrong writing system; Section~\ref{sec:script}).}
\label{tab:fmix}
\end{table}

\paragraph{Quiet English.} Against the script reference, extra words are the largest part of the error for nine of the thirteen rows: Gemini, for example, has 3.7\% extra words against 2.1\% wrong ones, and ElevenLabs 8.1\% against 2.8\%. Many of these extra words were really said: they are repeats and self-corrections the reader added. Against the spoken reference, extra words are the largest part for only four rows (Gemini, AAI 3.5, Zoom and ElevenLabs): Gemini's extra words fall from 3.7\% to 1.6\%, but ElevenLabs' only to 5.7\%, so most of ElevenLabs' extra words were not said. The mix also differs between systems with the same WER. Whisper SH and AAI 3.6 (8.3\% and 8.2\%) differ by only 0.1 points in WER, but Whisper SH misses more words (by 0.9 points, 95\% CI 0.1 to 2.4) and adds fewer (by 1.1 points, 95\% CI 0.5 to 1.8). Missing words are between 1.1\% and 2.7\% for every system.

\paragraph{Fan noise and other languages.} On the fan-on set, wrong words are the largest part for all thirteen rows; Deepgram has the most, 7.0\%. Outside English, wrong words are also the largest part for every system: 7.6--13.1\% of words, against 1.9--5.7\% missing and 0.5--1.2\% extra.

\paragraph{Types of substitution.} For every system, ordinary words are the largest group of wrong English words (46--63\%). Numbers written as words instead of digits or the other way round (Section~\ref{sec:numbers}), and near-misses such as ``request'' for ``requests'', together make up 33--50\%; names are 3--5\% (Appendix~\ref{app:extra}, Table~\ref{tab:fsubkind}).

\subsection{Whispered Speech}\label{sec:whisper}
\begin{figure}[!htbp]
\centering
\begin{tikzpicture}
\begin{axis}[xbar stacked, width=0.92\textwidth, height=7cm, bar width=8pt, xmin=0, xmax=52,
  symbolic y coords={Gemini Smart,Gemini,ElevenLabs NV,GPT-Live,RT Whisper,ElevenLabs,Whisper SH,Meta,Deepgram,AAI 3.6,AAI 3.5,Zoom,Qwen}, ytick=data, y dir=reverse, yticklabel style={font=\footnotesize},
  xlabel={Share of words (\%), whispered English}, xlabel style={font=\footnotesize}, xticklabel style={font=\footnotesize},
  axis y line*=left, axis x line*=bottom, xmajorgrids, major grid style={black!12},
  legend style={draw=none, font=\footnotesize, legend columns=3, at={(0.98,0.98)}, anchor=north east}, legend cell align=left]
\addplot[fill=figblue, draw=white] coordinates {(1.6,{Gemini Smart}) (2.9,{Gemini}) (4.0,{ElevenLabs NV}) (6.0,{GPT-Live}) (5.7,{RT Whisper}) (4.5,{ElevenLabs}) (6.0,{Whisper SH}) (7.1,{Meta}) (11.8,{Deepgram}) (3.0,{AAI 3.6}) (3.1,{AAI 3.5}) (5.1,{Zoom}) (11.2,{Qwen})};
\addplot[fill=figorange, draw=white] coordinates {(1.9,{Gemini Smart}) (1.0,{Gemini}) (1.7,{ElevenLabs NV}) (1.2,{GPT-Live}) (1.9,{RT Whisper}) (0.8,{ElevenLabs}) (1.8,{Whisper SH}) (1.1,{Meta}) (2.7,{Deepgram}) (25.7,{AAI 3.6}) (25.7,{AAI 3.5}) (27.6,{Zoom}) (32.2,{Qwen})};
\addplot[fill=figaqua, draw=white] coordinates {(0.7,{Gemini Smart}) (1.1,{Gemini}) (1.4,{ElevenLabs NV}) (1.7,{GPT-Live}) (1.3,{RT Whisper}) (3.8,{ElevenLabs}) (2.3,{Whisper SH}) (3.1,{Meta}) (5.3,{Deepgram}) (1.8,{AAI 3.6}) (2.2,{AAI 3.5}) (1.9,{Zoom}) (3.3,{Qwen})};
\legend{Wrong, Missing, Extra}
\node[anchor=west, font=\scriptsize, text=black!65] at (axis cs:4.2,{Gemini Smart}) {4.2};
\node[anchor=west, font=\scriptsize, text=black!65] at (axis cs:5.0,{Gemini}) {5.0};
\node[anchor=west, font=\scriptsize, text=black!65] at (axis cs:7.1,{ElevenLabs NV}) {7.1};
\node[anchor=west, font=\scriptsize, text=black!65] at (axis cs:8.9,{GPT-Live}) {8.8};
\node[anchor=west, font=\scriptsize, text=black!65] at (axis cs:8.9,{RT Whisper}) {8.9};
\node[anchor=west, font=\scriptsize, text=black!65] at (axis cs:9.1,{ElevenLabs}) {9.1};
\node[anchor=west, font=\scriptsize, text=black!65] at (axis cs:10.1,{Whisper SH}) {10.1};
\node[anchor=west, font=\scriptsize, text=black!65] at (axis cs:11.299999999999999,{Meta}) {11.3};
\node[anchor=west, font=\scriptsize, text=black!65] at (axis cs:19.8,{Deepgram}) {19.8};
\node[anchor=west, font=\scriptsize, text=black!65] at (axis cs:30.5,{AAI 3.6}) {30.5};
\node[anchor=west, font=\scriptsize, text=black!65] at (axis cs:31.0,{AAI 3.5}) {30.9};
\node[anchor=west, font=\scriptsize, text=black!65] at (axis cs:34.6,{Zoom}) {34.7};
\node[anchor=west, font=\scriptsize, text=black!65] at (axis cs:46.7,{Qwen}) {46.6};
\end{axis}
\end{tikzpicture}
\caption{Whispered WER split into wrong, missing and extra words, as a share of all words; the number is the WER.}
\label{fig:whispermix}
\end{figure}
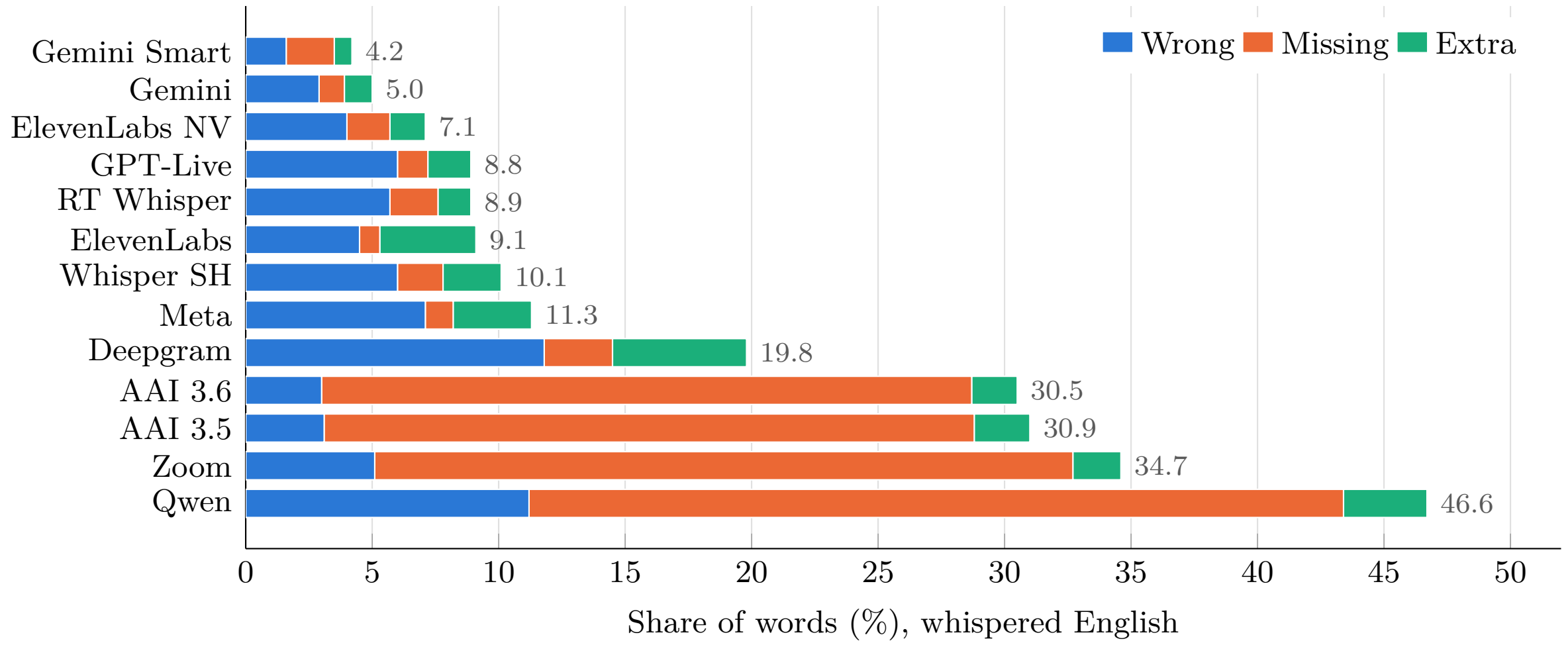

\paragraph{Four systems leave out whispered words.} On whispered speech, Qwen, Zoom and both AAI models leave out 26--32\% of words; every other row leaves out at most 3\% (Figure~\ref{fig:whispermix}). With 16 clips these shares are uncertain (for Qwen, 95\% CI 15--58\%), but each of the four intervals lies above 3\% (this also holds with bias-corrected intervals and after a Holm correction). Their errors are concentrated in some clips: 5 or 6 of the 16 whispered clips fail (WER of 50\% or more), and 2--4 are almost entirely lost (90\% or more). In the other group, no whispered clip fails, and wrong words are the largest part of the error for every row except Gemini Smart, whose few errors are split about evenly between wrong and missing words.

\paragraph{Position of deletions.} Both AAI models cut off the last 3 or more words of 4 of the 16 whispered clips. Qwen's missing stretches appear anywhere, including at the start. A cut-off ending leaves no visible gap in the text, and WER does not show it. AAI 3.6 (30.5\%) and Zoom (34.7\%) have similar whispered WER, but 4 of AAI 3.6's whispered clips have a cut-off ending and none of Zoom's (too few clips to test).

\paragraph{Recognition, turn detection or speech detection?} Qwen, Zoom and the AAI models all decide on the server when speech or a turn has ended, so a missing whispered word could be lost there rather than in recognition. This was tested on the 16 whispered clips (Appendix~\ref{app:systems}). For AAI, forcing the end of the last turn before closing the session changed nothing (AAI 3.6: 25.7\% of words missing, then 25.6\%), and turning end-of-turn detection off left 22.7\% missing for AAI 3.6 and 25.7\% for AAI 3.5. For Zoom, ending a segment after 5\,s of silence instead of 600\,ms lowered missing words from 27.6\% to 21.2\%. So AAI and, for the most part, Zoom lose these words whatever their turn settings; since the inside of a hosted service is not visible, this is counted as recognition, which here includes any speech detection inside the service. Qwen is different: our deployment runs its own speech detection before the model, with default settings, and relaxing it lowered Qwen's missing words from 32.2\% to 5.7\%. Qwen's whispered loss is therefore in the audio path of our own deployment, the level before the model, as for gated Whisper (Section~\ref{sec:whisperguide}). The relaxed setting did not raise WER on quiet or fan-on English or write text on silence (Appendix~\ref{app:systems}); the tables keep the deployment's default.

\paragraph{Replication on EARS.} EARS \citep{richter2024ears} has 107 speakers recorded in a sound-proofed studio, reading the same passage (the Rainbow Passage) normally and in a whisper; parts 1--4 from every speaker are used, 428 normal and 428 whispered clips. Table~\ref{tab:ears} gives the results, scored as in Appendix~\ref{app:ext}. Normal speech is easy for every system here (0.8--2.5\%). Every speaker reads the same well-known passage, so EARS tests the same text across many voices, not unseen text (Section~\ref{sec:limits}). Whispered, the same four systems again lose words, now 7--31\% of them: AAI 3.6 and AAI 3.5 rise by about 7 points, Zoom by 24 and Qwen by 31 (with its deployment's default speech detection, as in the tables). Whisper SH, which lost few words on our whispered clips, loses 10.4\% of words as missing on EARS, against 7.3\% for each AAI model (within chance), so the two groups of our in-house test are not a clean split across speakers. Gemini (in both settings), ElevenLabs (in both settings), Meta, GPT-Live and RT Whisper change by less than one point, and no speaker reaches a whispered WER of 20\% with any of them.

\begin{table}[!htbp]
\centering\small
\begin{adjustbox}{max width=\textwidth}
\begin{tabular}{@{}lrrrrr@{}}
\toprule
& & & Change & Whispered: & Speakers \\
System & Normal & Whispered & (paired) & wrong / missing / extra & $\geq$20\% \\
\midrule
Whisper SH & 1.5 & 11.5 {\scriptsize [8.2, 15.4]} & +10.0 {\scriptsize [+6.5, +14.0]} & 0.8 / 10.4 / 0.3 & 20 \\
Gated Whisper & 1.0 & 42.9 {\scriptsize [36.7, 49.4]} & +41.9 {\scriptsize [+35.1, +48.4]} & 3.1 / 39.1 / 0.7 & 67 \\
Qwen & 1.1 & 32.0 {\scriptsize [26.3, 38.1]} & +30.9 {\scriptsize [+25.0, +37.0]} & 1.0 / 30.6 / 0.4 & 56 \\
Meta & 1.4 & 1.7 {\scriptsize [1.3, 2.1]} & +0.3 {\scriptsize [0.0, +0.6]} & 0.8 / 0.3 / 0.5 & 0 \\
Gemini & 0.9 & 1.1 {\scriptsize [0.9, 1.4]} & +0.2 {\scriptsize [+0.0, +0.5]} & 0.5 / 0.2 / 0.3 & 0 \\
Gemini Smart & 0.8 & 1.0 {\scriptsize [0.8, 1.3]} & +0.2 {\scriptsize [0.0, +0.5]} & 0.6 / 0.2 / 0.2 & 0 \\
Deepgram & 2.5 & 4.0 {\scriptsize [3.2, 4.8]} & +1.5 {\scriptsize [+0.8, +2.4]} & 1.8 / 1.5 / 0.6 & 1 \\
AAI 3.6 & 1.2 & 8.1 {\scriptsize [5.2, 11.3]} & +6.9 {\scriptsize [+4.1, +10.2]} & 0.5 / 7.3 / 0.3 & 17 \\
AAI 3.5 & 1.2 & 8.3 {\scriptsize [5.6, 11.6]} & +7.1 {\scriptsize [+4.3, +10.4]} & 0.6 / 7.3 / 0.4 & 16 \\
Zoom & 0.9 & 25.0 {\scriptsize [20.1, 30.6]} & +24.1 {\scriptsize [+19.0, +29.9]} & 1.3 / 23.3 / 0.4 & 45 \\
ElevenLabs & 1.3 & 1.5 {\scriptsize [1.1, 1.8]} & +0.1 {\scriptsize [$-$0.2, +0.5]} & 0.7 / 0.2 / 0.5 & 0 \\
ElevenLabs NV & 1.1 & 1.2 {\scriptsize [0.9, 1.6]} & +0.2 {\scriptsize [$-$0.2, +0.5]} & 0.7 / 0.2 / 0.3 & 0 \\
GPT-Live & 1.3 & 2.1 {\scriptsize [1.7, 2.5]} & +0.8 {\scriptsize [+0.4, +1.2]} & 1.2 / 0.4 / 0.4 & 0 \\
RT Whisper & 1.2 & 2.1 {\scriptsize [1.7, 2.5]} & +0.9 {\scriptsize [+0.5, +1.3]} & 1.2 / 0.4 / 0.4 & 0 \\
\bottomrule
\end{tabular}
\end{adjustbox}
\caption{EARS: the same 107 speakers reading the same passage normally and whispered. WER (\%), numbers and spellings normalised; 95\% intervals from resampling speakers. Change: whispered minus normal, in points, paired by speaker and passage. Speakers $\geq$20\%: speakers whose whispered WER is 20\% or more. Gated Whisper: gated Whisper with its VAD dropping audio (Section~\ref{sec:whisperguide}).}
\label{tab:ears}
\end{table}

\subsection{Named Entities}\label{sec:names}
\paragraph{Names are written correctly less often than other words.} Our English scripts name people, products and apps (for example, ``send it to Maria'' or ``post it in Notion''). Over all English dictation, every system writes names correctly less often than other words (Table~\ref{tab:fnames}). Deepgram, used without its custom-vocabulary option, has the largest gap: 9.7 points (95\% CI 3.3 to 16.7). Five more systems have gaps of 4.8--6.5 points with intervals that exclude zero, but these are 13 comparisons, and after a Holm correction only Deepgram's gap remains clear. Names are 183 of 10{,}776 words, so even Deepgram's gap adds only about 0.2 points to its WER: a system can write many names wrongly and still have a low WER. Much of the gap comes from fan-on and whispered messages. On quiet English alone (123 name words), no system's gap is clearly above zero: Deepgram's is the largest (6.4 points, 95\% CI $-0.4$ to 14.7), and Whisper SH's and Qwen's fall from 6.5 to under 2 points.

\paragraph{Identifying names.} Of the 327 words capitalised in the middle of a sentence, only 156 are names, so the capital-letter rule alone finds names poorly; a written rule gives the 183 name words that are scored (Appendix~\ref{app:data}).

\begin{table}[!htbp]
\centering\small
\begin{adjustbox}{max width=\textwidth}
\begin{tabular}{@{}lrrrr@{}}
\toprule
System & Names & Other words & Gap & 95\% CI \\
\midrule
Whisper SH & 88.0 & 94.5 & 6.5 & [1.1, 12.9] \\
Qwen & 82.0 & 88.4 & 6.5 & [0.1, 14.1] \\
Meta & 90.2 & 95.0 & 4.8 & [0.3, 10.2] \\
Gemini & 93.4 & 96.9 & 3.5 & [$-$0.3, 8.2] \\
Gemini Smart & 95.1 & 96.8 & 1.7 & [$-$1.4, 5.6] \\
Deepgram & 81.4 & 91.1 & 9.7 & [3.3, 16.7] \\
AAI 3.6 & 92.3 & 93.0 & 0.7 & [$-$4.2, 6.0] \\
AAI 3.5 & 92.3 & 92.9 & 0.6 & [$-$4.2, 6.2] \\
Zoom & 90.7 & 92.7 & 2.0 & [$-$3.3, 8.0] \\
ElevenLabs & 91.8 & 95.4 & 3.6 & [$-$0.4, 8.5] \\
ElevenLabs NV & 94.5 & 96.1 & 1.6 & [$-$1.8, 5.6] \\
GPT-Live & 90.2 & 95.7 & 5.6 & [1.5, 10.8] \\
RT Whisper & 90.7 & 95.6 & 4.9 & [0.6, 10.3] \\
\bottomrule
\end{tabular}
\end{adjustbox}
\caption{Share of words written correctly (\%) for the 183 name words and for all other words in English dictation, and the gap in percentage points, with its 95\% confidence interval (bootstrap over the 106 messages).}
\label{tab:fnames}
\end{table}

\subsection{Dialects and Shared Errors}\label{sec:dialects}
\paragraph{Most failed clips are in four sets.} In the eleven other language sets, WER is below 5\% on several European sets and near or above 50\% on Emirati Arabic and, for some systems, Urdu. Almost all failed clips are in four sets: Saudi Arabic, Emirati Arabic, Hindi and Urdu (Appendix~\ref{app:extra}, Table~\ref{tab:fworst}). Urdu and Hindi are about writing (Section~\ref{sec:script}); the Arabic dialects are about recognition.

\paragraph{Arabic dialects.} Saudi and Emirati Arabic have failed clips for every one of the eleven systems: 15--35\% and 50--75\% of clips. On Emirati Arabic, WER is between 48.8\% and 61.8\% with default settings. Part of the error may come from the correct text itself: Arabic dialects have no standard spelling \citep{habash2012coda}, so a correct transcript can be spelled differently from ours.

\paragraph{Shared errors.} If every system fails on a clip, switching systems will not help; if only one fails, another system avoids it. Of the 220 non-English clips, 21 are hard clips (WER above 30\%) for all eleven systems: 14 Emirati Arabic, 6 Saudi Arabic and 1 Spanish (Spain). Across all 326 speech clips, 23 are hard for exactly one system (Appendix~\ref{app:extra}). In English, 67 of 73 quiet clips and all 17 fan-on clips are below 30\% for every system, while 13 of 16 whispered clips are above it for at least one: for English, the choice of system matters most on whispered speech.

\subsection{Far-Field and Noisy Speech}\label{sec:voices}
\paragraph{Far-field, babble and television noise.} VOiCES \citep{richey2018voices} replays audiobook sentences in a room and records them with a far microphone, in quiet, with several people talking (babble) and with a television playing. With babble, WER rises for every system, to between 38.7\% and 94.9\%, mostly as missing words; Zoom is lowest with babble and with television, although it is worse than average with a far microphone in a quiet room. Gemini returned no text within 30 seconds on 67 of the 100 babble clips, which count as empty transcripts; on the other 33 its WER is 40.3\%, so most of its babble WER is a failure to answer rather than misrecognition. Appendix~\ref{app:ext} gives the full results.

\FloatBarrier
\section{Output Formatting Errors}\label{sec:formatting}
\subsection{Number Formatting}\label{sec:numbers}
\paragraph{System-specific number habits.} Most benchmarks convert numbers to a common form before scoring; here they are not converted, so each system's ITN output is visible. The English scripts contain 276 numbers written in digits and 193 written as words (such as ``first'' or ``twenty''). Qwen writes 68\% of the digit numbers as words (``seven'' for ``7''), Meta 47\% and GPT-Live 45\%; the two AAI models do the opposite, turning 41--42\% of number words into digits; Whisper SH, Gemini, Zoom and ElevenLabs usually keep the form in the script (Appendix~\ref{app:extra}, Table~\ref{tab:fnumbers}). These switches make up 16.1\% of GPT-Live's English errors, 12.9\% of Meta's and 10.4\% of Qwen's, against 4.8--6.1\% for ElevenLabs, Zoom and Whisper SH; a number written in several words counts once, so these shares are a minimum. The number itself was heard correctly.

\paragraph{Effect of normalisation on ranking.} Re-scored after Whisper's English text normaliser, which writes numbers as digits and removes fillers, every quiet-English WER falls, by 1.0--3.8 points when both scores use the same word splitting, but the order of systems changes little (rank correlation 0.98; Appendix~\ref{app:extra}, Table~\ref{tab:normwer}). Number habits matter to a user who sees the text, but they are not what separates the systems on quiet English.

\subsection{Writing System}\label{sec:script}
\paragraph{Urdu.} Urdu shows the biggest gap between systems, and the cause is mostly the writing system. Table~\ref{tab:urdu} counts, for each system that writes Urdu in another script, the clips written mostly in Devanagari, the script of Hindi. Qwen, Meta and AAI 3.5 write most Urdu clips in Devanagari and fail on 85--95\% of them. Meta and AAI 3.5 have no Urdu option and detect the language themselves; Qwen does this even when Urdu is selected. Spoken Urdu and Hindi are close, so a system detecting the language itself may reasonably hear Hindi; but a user who chose Urdu receives text in the wrong writing system, whatever words it contains. Zoom, which also has no Urdu option, writes 19 of 20 Urdu clips in Latin letters (``Ek bum governoral journal ke daftar\ldots'') and fails on every one. Both Whisper rows, Deepgram, AAI 3.6 and ElevenLabs write in the Urdu writing system and fail on at most 5\% of clips; RT Whisper has the lowest Urdu WER (12.3\%).

\begin{table}[!htbp]
\centering\small
\begin{tabular}{@{}llrr@{}}
\toprule
System & Urdu option & Mostly Devanagari & Failed (WER $\geq$50\%) \\
\midrule
Qwen & selected & 17 & 19 \\
AAI 3.5 & none & 18 & 18 \\
Meta & none & 17 & 17 \\
Gemini & none & 9 & 9 \\
Gemini Smart & none & 4 & 5 \\
GPT-Live & selected & 1 & 1 \\
\bottomrule
\end{tabular}
\caption{Urdu clips (of 20) written mostly in Devanagari, and Urdu clips with a WER of 50\% or more. Gemini Smart writes some Devanagari on a fifth clip, and GPT-Live on 3 clips in all. Zoom writes 19 of 20 in Latin letters.}
\label{tab:urdu}
\end{table}

\paragraph{Hindi.} Zoom has the highest Hindi WER of the eleven systems (34.6\%), with text in the right writing system. Its CER (16.3\%) is less than half its WER, so many of its wrong words differ from the correct word by a few characters, for example in vowel signs, rather than being different words (Appendix~\ref{app:extra}).

\subsection{Disfluency Removal and Reference Choice}\label{sec:cleanup}
\paragraph{Effect of the reference.} The script reference asks whether the text is the message the user meant; the spoken reference asks whether the system heard what was said. A clean-up setting removes fillers, repeats and self-corrections on purpose, so it scores well on the first and loses points on the second. In the literature this is disfluency removal, and the choice between verbatim and clean transcripts \citep{jamshidlou2020disfluency}.

\paragraph{Scored against the script.} Smart lowers Gemini's quiet-English WER from 6.9\% to 3.9\%, and \texttt{no\_verbatim} lowers ElevenLabs' from 12.8\% to 5.1\%, in both cases almost all through fewer extra words (3.7\% to 0.7\%, and 8.1\% to 1.4\%).

\begin{table}[!htbp]
\centering\small
\begin{adjustbox}{max width=\textwidth}
\begin{tabular}{@{}lrrrrrr@{}}
\toprule
& \multicolumn{2}{c}{Quiet} & \multicolumn{2}{c}{Fan on} & \multicolumn{2}{c}{Whispered} \\
\cmidrule(lr){2-3}\cmidrule(lr){4-5}\cmidrule(l){6-7}
System & Script & Spoken & Script & Spoken & Script & Spoken \\
\midrule
Whisper SH & 8.3 & 6.1 {\scriptsize [4.9, 7.7]} & 6.5 & 4.6 {\scriptsize [2.8, 6.8]} & 10.1 & 9.3 {\scriptsize [6.2, 12.5]} \\
Qwen & 13.6 & 9.8 {\scriptsize [7.2, 12.9]} & 7.0 & 5.1 {\scriptsize [3.7, 6.6]} & 46.6 & 45.5 {\scriptsize [31.6, 66.5]} \\
Meta & 9.5 & 5.5 {\scriptsize [4.7, 6.4]} & 7.4 & 4.8 {\scriptsize [3.1, 6.8]} & 11.3 & 9.9 {\scriptsize [7.7, 12.0]} \\
Gemini & 6.9 & 3.7 {\scriptsize [3.0, 4.4]} & 2.5 & 1.9 {\scriptsize [0.8, 3.1]} & 5.0 & 4.6 {\scriptsize [2.4, 6.1]} \\
Gemini Smart & 3.9 & 4.6 {\scriptsize [3.8, 5.5]} & 3.2 & 3.3 {\scriptsize [1.7, 5.3]} & 4.2 & 4.4 {\scriptsize [2.9, 6.4]} \\
Deepgram & 13.6 & 9.6 {\scriptsize [8.3, 10.8]} & 11.4 & 9.3 {\scriptsize [6.8, 11.6]} & 19.8 & 18.3 {\scriptsize [14.3, 22.0]} \\
AAI 3.6 & 8.2 & 4.8 {\scriptsize [4.0, 5.6]} & 4.1 & 1.7 {\scriptsize [1.1, 2.2]} & 30.5 & 29.5 {\scriptsize [13.1, 54.4]} \\
AAI 3.5 & 9.0 & 5.6 {\scriptsize [4.8, 6.4]} & 4.3 & 2.0 {\scriptsize [1.3, 2.5]} & 30.9 & 29.9 {\scriptsize [13.6, 55.7]} \\
Zoom & 7.4 & 4.8 {\scriptsize [4.1, 5.6]} & 3.6 & 2.3 {\scriptsize [1.1, 3.8]} & 34.7 & 33.5 {\scriptsize [17.0, 59.6]} \\
ElevenLabs & 12.8 & 8.8 {\scriptsize [7.1, 10.8]} & 5.8 & 3.3 {\scriptsize [2.0, 4.8]} & 9.1 & 7.6 {\scriptsize [4.1, 10.7]} \\
ElevenLabs NV & 5.1 & 4.5 {\scriptsize [3.8, 5.4]} & 4.5 & 2.5 {\scriptsize [1.4, 3.8]} & 7.1 & 6.6 {\scriptsize [2.6, 10.8]} \\
GPT-Live & 7.3 & 5.0 {\scriptsize [4.2, 5.9]} & 3.6 & 3.6 {\scriptsize [2.1, 5.4]} & 8.8 & 8.1 {\scriptsize [6.3, 9.5]} \\
RT Whisper & 8.2 & 5.0 {\scriptsize [4.1, 5.9]} & 4.6 & 3.7 {\scriptsize [1.9, 5.8]} & 8.9 & 8.4 {\scriptsize [6.7, 10.4]} \\
\bottomrule
\end{tabular}
\end{adjustbox}
\caption{English WER (\%) against the script reference (as in Table~\ref{tab:overall}) and against the spoken reference, with its 95\% confidence interval (bootstrap over messages). Lower is better.}
\label{tab:verbatim}
\end{table}

\paragraph{Scored against the spoken reference.} Table~\ref{tab:verbatim} scores the same transcripts against the spoken reference. Every system run at its verbatim default does better on quiet English, by 2.2--4.0 points, because words the reader added are no longer counted as extra. Gemini Smart is the only row that does worse (+0.8 points, paired 95\% CI $-0.1$ to 1.6): it leaves out the repeats and self-corrections that are now in the reference. On quiet English, Gemini (3.7\%), ElevenLabs NV (4.5\%) and Gemini Smart (4.6\%) are now close. In paired comparisons, Gemini is 0.9 points below Gemini Smart (95\% CI $-1.8$ to 0.0), and Gemini Smart and ElevenLabs NV are 0.1 points apart. Against the script, Gemini Smart is clearly below ElevenLabs NV (by 1.2 points, 95\% CI 0.6 to 1.9); against what was said, no clean-up setting is clearly ahead. The whispered results of Section~\ref{sec:whisper} hold against either reference: every whispered WER moves by at most 1.5 points.

\paragraph{Dependence of the spoken reference on the systems.} The spoken reference was drafted from where the verbatim systems agreed, so it could favour them. Each system voted for 237 to 313 of the 378 accepted edits. The reference was rebuilt eleven times, each time keeping only the accepted edits that the other ten systems would still have proposed. No system's WER against its own rebuilt reference rose by more than 0.3 points in any condition. With the stricter thresholds used to pre-tick proposals, the largest rise was 1.2 points, for Meta on quiet English. But the order of Gemini, Gemini Smart and ElevenLabs NV is not stable. Without Gemini's votes, Gemini is 0.5 points below Gemini Smart (95\% CI $-1.3$ to 0.3); with the stricter pre-tick thresholds, it is 0.8 points above (95\% CI $-0.1$ to 1.7). Neither difference is clear. A clean-up setting is clearly best only against the script.

\FloatBarrier
\section{Language Handling in Code-Switched Recordings}\label{sec:cs}
\paragraph{With only English selected, some systems remove the other language.} The English-only setting is called a \emph{filter} for a system when it loses the other language in more than half of the EN\,+\,X recordings, while the both-languages setting loses it in fewer than half. It is called a \emph{preference} when the English-only setting itself loses the other language in fewer than half. By this rule, the setting is a filter for Gemini (in both settings), Deepgram and Whisper SH. They lose the non-English part in 86--99\% of the 100 recordings with only English selected, against 33--46\% with both selected (Table~\ref{tab:fcs}, first column; Table~\ref{tab:csoverall} counts a loss of either part). It is a preference for Qwen, Meta, AAI 3.6, AAI 3.5, GPT-Live and RT Whisper, which lose it in 7--38\%. Zoom and ElevenLabs (in both settings) lose most second languages whatever is selected, for reasons given below. Whether a filter is right depends on the app; a user who selects English and then says a sentence in another language may or may not want it written.

\paragraph{Recording construction.} Our recordings are not natural code-switching \citep{poplack1980sometimes,winata2023decades}: they join single-language recordings from different speakers, with a 300\,ms pause between them, so the language, the speaker and the recording all change at a sentence boundary. One hundred recordings pair English with one of the 10 other language sets (EN\,+\,X), and 300 pair two non-English languages (X\,+\,Y). Urdu is left out because several systems write it in the wrong writing system even alone (Section~\ref{sec:script}). English comes first in half of the EN\,+\,X recordings, and the order of each X\,+\,Y pair is balanced likewise. Each recording is played with both languages selected and with automatic detection, and EN\,+\,X recordings also with only English selected. All eleven systems were run on this test.

To score a recording, the transcript is lined up with the two parts' correct texts joined end to end, and each reference word is marked as heard, replaced or missing. A part is \emph{missing} when less than 20\% of its words are heard: either nothing is written in its place, or other text is. Table~\ref{tab:csoverall} gives the share of recordings in which a part goes missing. A part may also go missing because it is hard on its own; on a screened subset of recordings whose parts every system gets mostly right alone, the order of the systems is almost the same and the conclusions do not change (Appendix~\ref{app:cs}).

\begin{table}[!htbp]
\centering\small
\begin{adjustbox}{max width=\textwidth}
\begin{tabular}{@{}lrrrrr@{}}
\toprule
& \multicolumn{3}{c}{EN\,+\,X (100 recordings)} & \multicolumn{2}{c}{X\,+\,Y (300 recordings)} \\
\cmidrule(lr){2-4}\cmidrule(l){5-6}
System & Both selected & English only & Auto-detect & Both selected & Auto-detect \\
\midrule
Whisper SH & 45 & 87 & 49 & 87 & 88 \\
Qwen & 20 & 38 & 20 & 46 & 46 \\
Meta & 19 & 23 & 21 & 32 & 35 \\
Gemini & 50 & 93 & 71 & 36 & 75 \\
Gemini Smart & 52 & 87 & 73 & 62 & 83 \\
Deepgram & 40 & 99 & 40 & 58 & 58 \\
AAI 3.6 & 10 & 25 & 15 & 28 & 28 \\
AAI 3.5 & 13 & 35 & 16 & 23 & 33 \\
Zoom & 62 & 62 & 62 & 85 & 85 \\
ElevenLabs & 62 & 85 & 62 & 93 & 93 \\
ElevenLabs NV & 80 & 93 & 79 & 96 & 97 \\
GPT-Live & \textbf{6} & \textbf{7} & \textbf{7} & \textbf{4} & \textbf{5} \\
RT Whisper & 11 & 12 & 10 & 11 & 11 \\
\bottomrule
\end{tabular}
\end{adjustbox}
\caption{Share of two-language recordings (\%) in which one of the two languages went missing; lowest in each column in bold. EN\,+\,X: 100 recordings of English and another language; X\,+\,Y: 300 recordings of two other languages. For shares between 25\% and 75\%, the 95\% interval is about $\pm$12--22 points for EN\,+\,X and $\pm$8--11 points for X\,+\,Y, wide because recordings reuse source clips (Appendix~\ref{app:data}).}
\label{tab:csoverall}
\end{table}

\paragraph{Which part is lost.} Table~\ref{tab:fcs} (Appendix~\ref{app:cs}) shows which part goes missing and what appears in its place. Meta never loses the first part, in any setting; all of its losses are the second part. Gemini does the opposite: with automatic detection, it loses the first part in 54\% of EN\,+\,X recordings and the second in 19\%, and the lost part is the English one in 37 of the 100. GPT-Live loses a part in 6--7\% of EN\,+\,X recordings, almost always the second. Qwen also loses the second part more often than the first (14\% against 6\% of EN\,+\,X recordings with both languages selected, and 101 against 37 of the 300 X\,+\,Y); in 14 of its 20 EN\,+\,X losses it writes other text in the lost part's place. Whisper SH usually leaves a gap: with both languages selected, it loses the first part in 25\% of recordings and the second in 20\%, and in 43\% nothing is written where the part should be.

\paragraph{Romanised and translated parts leave no gap.} Deepgram rarely leaves a gap. Instead, in 34\% of the 50 EN\,+\,X recordings whose other language has its own writing system (Arabic, Hindi or Mandarin), it writes that part in Latin letters, mostly Arabic and Mandarin parts as strings of similar-sounding Latin syllables. Gemini, with both languages selected, writes the Hindi part in Latin letters in 5 of the 10 recordings of English and Hindi. A real English translation is mainly a Whisper behaviour, as reported for code-switched speech \citep{peng2023prompting}: Whisper SH, with only English selected, translates the other part into English in 18\% of these recordings. A check for missing text does not find romanised or translated parts.

\paragraph{Single-language sessions and late finalisation.} Zoom took one language per session, so all three settings behave like English only: it loses a language in 62\% of EN\,+\,X recordings and 85\% of X\,+\,Y. Its Arabic, Hindi and Mandarin parts are lost in every recording, while most Spanish, German and French parts survive (43 of 130, 7 of 110 and 19 of 110 lost, with both languages selected). ElevenLabs, by default, finalises the whole recording as one block of text. It loses a part in 62\% of EN\,+\,X recordings (almost always the non-English part) and 93\% of X\,+\,Y; ElevenLabs NV in 80\% and 96\%. Set to finalise at each 300\,ms pause, the gap placed between the two clips, ElevenLabs lost far fewer languages (Appendix~\ref{app:cs}).

\FloatBarrier
\section{Audio Handling Before the Model: A Whisper Case Study}\label{sec:whisperguide}
Whisper is a widely used open ASR model. Unlike the hosted services in this paper, it works on 30-second windows, so live use needs a growing buffer \citep{machacek2023whisperstreaming} or cutting at pauses \citep{bain2023whisperx}, and it leaves several choices to whoever runs it: where it is hosted, how the audio is cut into segments, and whether a VAD drops audio before the model hears it. On whispered speech, these choices changed our results more than the choice between most systems did. This section describes the setups tried and what each choice did in this test.

\paragraph{The nine setups.} Table~\ref{tab:setups} lists them. Setups 1--6 run Whisper Large V3 on two deployments of ours. OpenAI does not say which Whisper version it serves for setups 7--9; at its 2023 launch it described its Whisper API model as Large V2 \citep{openaiwhisperapi}, so differences between setups 1--6 and 7--9 may come from the model version as well as from how it is run. Appendix~\ref{app:whisper} gives the exact rules of each setup.

\begin{table}[!htbp]
\centering\small
\begin{adjustbox}{max width=\textwidth}
\begin{tabular}{@{}rlll@{}}
\toprule
& Setup & How the audio is cut into segments & Audio dropped before the model \\
\midrule
1 & Gated, VAD on (``gated Whisper'') & server's own speech detection & Silero VAD on the client drops silence \\
2 & Gated, VAD off & server's own speech detection & none \\
3 & Self-hosted, Silero cuts (``Whisper SH'') & client, at pauses Silero finds & none \\
4 & Self-hosted, Silero drops & client, at pauses Silero finds & Silero drops silence \\
5 & Self-hosted, TEN cuts & client, at pauses TEN VAD \citep{tenvad} finds & none \\
6 & Self-hosted, TEN drops & client, at pauses TEN finds & TEN drops silence \\
7 & RT Whisper & OpenAI's streaming service & none \\
8 & Whisper-1 & none: the whole recording is uploaded at the end & none \\
9 & Whisper-1 split & client, at pauses Silero finds & segments with no speech are not sent \\
\bottomrule
\end{tabular}
\end{adjustbox}
\caption{The nine ways of running Whisper. Setups 1--2 and 3--6 are two different deployments of Whisper Large V3; setups 7--9 are OpenAI's hosted Whisper.}
\label{tab:setups}
\end{table}

Table~\ref{tab:family} compares the nine on the same clips.

\begin{table}[!htbp]
\centering\small
\begin{adjustbox}{max width=\textwidth}
\begin{tabular}{@{}lrrrrrrrrr@{}}
\toprule
& \shortstack[r]{Gated\\VAD on} & \shortstack[r]{Gated\\VAD off} & \shortstack[r]{Self-hosted\\Silero cuts} & \shortstack[r]{Self-hosted\\Silero drops} & \shortstack[r]{Self-hosted\\TEN cuts} & \shortstack[r]{Self-hosted\\TEN drops} & \shortstack[r]{OpenAI\\RT Whisper} & \shortstack[r]{OpenAI\\Whisper-1} & \shortstack[r]{Whisper-1\\split} \\
\midrule
English, quiet & 8.7 & 9.5 & 8.3 & 8.0 & 7.8 & 7.2 & 8.2 & \textbf{5.5} & 6.5 \\
English, fan on & 7.0 & 7.5 & 6.5 & 5.3 & 5.9 & 5.9 & \textbf{4.6} & 5.3 & 6.3 \\
English, whispered & 80.7 & 61.1 & 10.1 & 51.4 & 10.2 & 10.7 & 8.9 & \textbf{7.7} & 23.4 \\
10 other languages$^\dagger$ & 17.7 & 17.3 & 18.4 & 18.6 & 18.4 & 17.6 & \textbf{14.0} & 18.8 & 18.9 \\
Hindi & 29.4 & 30.3 & 29.8 & 32.9 & 29.8 & 31.6 & \textbf{11.8} & 36.0 & 36.0 \\
Spanish (Spain) & 10.0 & 10.0 & 19.5 & 14.2 & 19.5 & 14.7 & \textbf{7.9} & 8.4 & 8.4 \\
English, quiet, formatted & 16.1 & 17.5 & 16.0 & 16.0 & 15.3 & 15.5 & 11.9 & \textbf{10.1} & 13.4 \\
\midrule
Wait, median (ms) & 395 & \textbf{360} & 361 & 385 & 390 & 395 & 735 & 1,850 & 1,550 \\
Wait, 90th percentile (ms) & 624 & 596 & \textbf{498} & 540 & 533 & 538 & 832 & 4,105 & 2,314 \\
No-speech clips with text (of 10) & \textbf{0} & \textbf{0} & 6 & \textbf{0} & 6 & 2 & \textbf{0} & 10 & \textbf{0} \\
\midrule
Lost a language, EN\,+\,X (\%) & 34 & -- & 45 & 53 & 39 & 54 & \textbf{11} & 89 & 68 \\
Lost a language, X\,+\,Y (\%) & 60 & -- & 87 & 96 & 93 & 96 & \textbf{11} & 94 & 91 \\
\bottomrule
\end{tabular}
\end{adjustbox}
\caption{Nine ways of running Whisper for dictation. Error rates in \%, lower is better; the lowest in each row is in bold. $^\dagger$Average over 10 language sets, not counting Urdu. Lost a language: share of all two-language recordings (100 EN\,+\,X and 300 X\,+\,Y) in which one language went missing, with both languages selected. Gated Whisper ran the two-language test with its VAD on only.}
\label{tab:family}
\end{table}

\paragraph{Finding 1: Silero VAD dropping audio lost whispered speech.} With its VAD on, the gated deployment has a whispered WER of 80.7\%, and the VAD never sent 2 of the 16 whispered clips. Turning the VAD off lowers whispered WER by 19.6 points (95\% CI 7.0 to 36.9) and changes quiet and fan-on English by less than a point (Appendix~\ref{app:whisper}, Table~\ref{tab:vad}). The self-hosted deployment shows the same effect from the other side: sent all of the audio, its whispered WER is 10.1\%; with Silero dropping audio, 51.4\%, with 44.6\% of words missing. Both effects hold across the 107 EARS speakers. Turning the gated deployment's VAD off lowers whispered WER by 10.1 points (95\% CI 8.0 to 12.2). Silero dropping audio raises the self-hosted deployment's by 26.8 points (95\% CI 21.3 to 32.2), and puts 62 of the 107 speakers at 20\% or more, against 20 (Table~\ref{tab:earsvad}).

\paragraph{Finding 2: TEN VAD dropped audio with a much smaller loss.} When the VAD only cuts the audio, Silero and TEN give nearly the same results (whispered 10.1\% and 10.2\%). When it also drops audio, they differ sharply: with Silero, whispered WER rises to 51.4\%; with TEN, it stays at 10.7\%. On EARS, dropping with Silero gives 38.3\% and dropping with TEN 14.4\%, against 11.5\% for sending all of the audio; TEN's rise of 2.9 points is within chance (95\% CI $-0.4$ to 6.7). TEN keeps whispered speech because it calls most of it speech: 79\% of whispered 32\,ms windows at its default threshold, against 37\% for Silero. It does not call steady noise speech any more than Silero does, but it calls more background talk speech (Appendix~\ref{app:whisper}).

\paragraph{Finding 3: lowering Silero's threshold did not fix it.} Even at a threshold of 0.2, Silero's whispered WER stays at 41.1\%, while TEN's stays at 10.0--10.7\% at every threshold tried (Appendix~\ref{app:whisper}, Figure~\ref{fig:vadsweep}).

\paragraph{Finding 4: Whisper wrote invented text on silence.} Whisper can write text that was never said (hallucination) \citep{koenecke2024careless,mansoor2025mitigating}. On all 10 no-speech clips Whisper-1 wrote text, mostly ``you'' and ``Thank you for watching.'', and the self-hosted deployment wrote a stray ``.'' or ``so'' on 6 of 10. On two short Spanish (Spain) clips the self-hosted deployment replaced the whole speech with a stock phrase such as ``Muchas gracias.'' Dropping audio with a VAD removes text on silence, at the cost above.

\paragraph{Finding 5: the most accurate setups were the slowest.} Whisper-1, which transcribes the whole recording after the user stops, has the lowest quiet and whispered English WER of the nine setups (5.5\% and 7.7\%), but a typical wait of 1{,}850\,ms and text on every no-speech clip. RT Whisper is the most accurate on the other languages (14.0\% average, against 17.3--18.9\%) and loses a language in 11\% of two-language recordings, but accepts only one language and waits about twice as long as the self-hosted deployment (735\,ms against 361\,ms typical).

\paragraph{Finding 6: cutting at the speaker's pauses kept whispered words.} With its VAD off, the gated deployment still has a whispered WER of 61.1\%, almost all of it missing words; the self-hosted deployment, which transcribes the segments the client cuts at pauses, misses 1.8\% of whispered words (10.4\% on EARS). The two are different deployments, so this does not isolate how the audio is cut (Appendix~\ref{app:whisper}).

\FloatBarrier
\section{Recommendations for Evaluating Dictation ASR}\label{sec:guidance}
Systems with similar error rates made different kinds of errors in our test, at every level of the taxonomy. Tests of ASR systems for dictation should include the following.
\begin{itemize}\itemsep1pt
  \item \textbf{Recognition: report wrong, missing and extra words separately}, with missing stretches and cut-off endings (Sections~\ref{sec:mix} and~\ref{sec:whisper}). Test whispered speech with several speakers. Report accuracy on names separately, because a gap of 1 point and one of 10 look almost the same in overall WER, and check which words are names (Section~\ref{sec:names}).
  \item \textbf{Output formatting: say which reference is used}, and score against both what the speaker meant and what they said when clean-up settings are compared (Section~\ref{sec:cleanup}). Check how numbers are written (Section~\ref{sec:numbers}) and the writing system of the output, especially for languages written in more than one (Section~\ref{sec:script}).
  \item \textbf{Language handling: test two-language speech under each language setting}, and tell missing, romanised and translated parts apart (Section~\ref{sec:cs}). Settings that seem unrelated to language can matter: ElevenLabs lost far fewer languages when it finalised text at each pause.
  \item \textbf{Before the model: test the whole app}, including any VAD and how the audio is cut, on whispered speech and on silence (Section~\ref{sec:whisperguide}).
  \item \textbf{Separate errors all systems share from errors one system makes.} Errors every system makes, such as on Saudi and Emirati Arabic, will not be fixed by switching among the tested systems (Section~\ref{sec:dialects}).
\end{itemize}

\FloatBarrier
\section{Limitations and Future Work}\label{sec:limits}
The study compares systems on the same audio under controlled conditions, and that design sets its scope. Each point below is also a direction that the released code and data support.
\begin{itemize}\itemsep1pt
  \item \textbf{Speakers and speech style.} The in-house English is 106 scripted messages from one speaker. This keeps the voice constant, so differences between systems and conditions are not confounded with the speaker; the reader varied accent deliberately. Read speech has fewer and different disfluencies than spontaneous dictation, which matters most for the clean-up results. The EARS and VOiCES results extend the main findings to 107 and 100 speakers; spontaneous dictation from many speakers, recorded and scored with the same pipeline, is the natural next step.
  \item \textbf{Set sizes.} With 16 whispered messages and 20 clips per language set, the full matrix of systems, settings and languages stays tractable. The tables are best read as patterns, with the intervals given; the differences called clear hold after correction for multiple comparisons. Larger sets per language would tighten the estimates for individual languages.
  \item \textbf{Corpora.} Each dialect set comes from a single public corpus, so dialect and recording source are not separated. The external corpora are read speech with public texts (EARS uses one well-known passage, VOiCES uses LibriSpeech \citep{panayotov2015librispeech}), so a system may have seen them in training, which could flatter clean scores and help on the whispered EARS passage. A second corpus per dialect and unseen whispered text would separate these effects.
  \item \textbf{References and labels.} The spoken reference keeps every departure from the script that the systems heard and a listener confirmed; a departure no system heard stays as scripted, and fillers are left out, as in the scripts. Most analyses use the script, which favours clean-up settings on extra words, so Section~\ref{sec:cleanup} reports both. Names were labelled by a language model (Claude) following a written rule, accents by a classifier checked against the reader's account, and romanised and translated parts by rules checked by hand on examples. The thresholds in Table~\ref{tab:taxonomy} are stated so that others can vary them, and a fully hand-annotated verbatim reference would extend the comparison.
  \item \textbf{Configuration.} Each system runs at its defaults, with the changes in Appendix~\ref{app:systems}, so the results describe systems as a developer first meets them. Where a setting could change a conclusion, it was tested directly: end-of-turn detection, regional language codes, GPT-Live's delay (Table~\ref{tab:gptdelay}), ElevenLabs' pause finalising and the nine Whisper setups. The Whisper setups in the main comparison were chosen using Section~\ref{sec:whisperguide}, and GPT-Live's delay on 50 English test clips, with both settings reported. Further settings to explore include ElevenLabs' pause finalising on single-language audio, gated Whisper's two-language runs without its VAD, Zoom settings beyond its sample app, and repeated runs of the VAD threshold sweep.
  \item \textbf{Streaming behaviour.} The finalised text is scored and the wait for it measured, which is what a dictation user receives. Partial text and its stability \citep{shangguan2020stability}, and endpointing decided by the system itself, are separate questions the released client can be extended to measure.
  \item \textbf{Hosted systems change.} Results reflect the hosted services in October 2026, tested from one client location. Each clip was played once per system and setting; a replay a week later gave near-identical results (Appendix~\ref{app:systems}). Because the client and scoring are released, the comparison can be rerun as services change.
\end{itemize}

\FloatBarrier
\section{Conclusion}\label{sec:conclusion}
The same audio was streamed to eleven ASR systems, and every error was grouped by the level it shows up at. Systems with similar error rates failed differently at every level. In recognition, some systems left out a large share of whispered words that the others heard, and the pattern largely held across 107 external speakers; names were written correctly less often than other words. In output formatting, systems differed in how they wrote numbers and in which writing system, and clean-up settings wrote the message meant rather than every word said. In language handling, selecting only English removed the other language for some systems and only guided it for others. Before the model, how the app detected speech and dropped audio decided much of the whispered-speech result, for Whisper and for our Qwen deployment. Some errors, such as those on Saudi and Emirati Arabic, are made by every system. These error types should be reported alongside WER, and the app's whole audio path should be tested.

\FloatBarrier
\section*{Ethics}
The English dictations were recorded by one member of our team, who volunteered to read scripted messages for this study. They consented to the use of their recordings for it and to their public release, and the consent is kept on record by the authors. The scripts were written for the study, and the people they name are invented, except that three email scripts are signed with the name of one of the authors, who agreed to this; the scripts contain no other personal data. No audio or text from users of any product is included. Because the test streams audio to hosted ASR services, the recordings were processed by those providers under their terms of service. The public clips are used under their licences. EARS (CC BY-NC 4.0) and VOiCES (CC BY 4.0) are used for non-commercial research evaluation only. No data from this study, and no transcript returned by a tested system, was used to train any model; the systems' transcripts were used only to score them and to draft the spoken references (Section~\ref{sec:setup}).

\paragraph{Code and data availability.} The test client, the scoring and analysis code, and the scripts that build every table in this paper are released at \url{https://github.com/heylemon-ai/streaming-dictation-asr} under the Apache 2.0 licence. The in-house English recordings (quiet, fan on and whispered, and the no-speech clips), both references for each, and every tested system's transcripts are released at \url{https://huggingface.co/datasets/hey-lemon/streaming-dictation-asr} under CC BY 4.0. The audio of the other language sets, the two-language recordings, EARS and VOiCES is not redistributed; the transcripts of those runs include the datasets' reference text, which remains under the original licences (EARS: CC BY-NC 4.0). The dataset lists the public clips used and the recipe for the two-language recordings; the repository has scripts that fetch EARS and VOiCES from their sources and rebuild the two-language recordings once the public clips are in place.

\paragraph{Competing interests.} The authors work at a company that makes dictation software and is developing its own ASR model, and that has commercial relationships with some of the tested providers. The authors and their company have no partnership with any of the tested providers. This work received no external funding: it was carried out and paid for by the authors' company, and no tested provider funded it.

\FloatBarrier
\appendix
\section{System Configuration}\label{app:systems}
Each system is used with its default settings, with these changes. Deepgram has \texttt{punctuate} and \texttt{smart\_format} on. GPT-Live uses \texttt{delay} \texttt{low}. Whisper SH decodes with beam width 5, is given both languages when two are selected, and has its audio cut into segments by the client (Section~\ref{sec:whisperguide}). ElevenLabs writes verbatim text and finalises once, when the audio ends. Zoom uses its sample app's settings with clean-up off; it has no locale for Egypt, Latin America, Puerto Rico or Urdu, so Saudi Arabic and Spain Spanish are used, and nothing for Urdu. A system with no option for a language is sent none.

\paragraph{GPT-Live's delay setting.} GPT-Live has no documented default \texttt{delay}, so both \texttt{low} (used in the main tables) and \texttt{minimal} were run on every test set (Table~\ref{tab:gptdelay}). Waits are the same (typical 744 and 752\,ms), and neither writes text on silence. \texttt{minimal} has higher WER on every English set and on 9 of the 11 other languages; after a Holm correction over the 25 comparisons, the difference is clear on quiet and whispered English, both EARS sets, three VOiCES conditions and Hindi.

\paragraph{Regional language codes.} In the main tables Gemini gets \texttt{ar-EG} for every Arabic set and \texttt{es-US} for every Spanish set, Zoom \texttt{ar-SA} and \texttt{es-ES}, and the others the bare language. With each provider's documented regional codes instead (Deepgram \texttt{ar-EG}, \texttt{ar-SA}, \texttt{ar-AE}, \texttt{es-419}; Zoom \texttt{ar-AE}; Gemini \texttt{es-419}, as it lists no Saudi or Emirati code). Transcripts were unchanged on 78 of 80 Deepgram clips, all 20 Zoom clips and 33 of 40 Gemini clips, and no failed clip changed; the largest WER change was 8.7\% to 7.8\% (Deepgram, Latin American Spanish).

\paragraph{End-of-turn and speech-detection checks.} The 16 whispered clips were also run with: AAI sending \texttt{ForceEndpoint} before \texttt{Terminate}; AAI with end-of-turn detection off (\texttt{end\_of\_turn\_\allowbreak confidence\_threshold} 1, \texttt{max\_turn\_\allowbreak silence} 10\,s); Zoom with \texttt{silence\_\allowbreak duration\_ms} 5{,}000 instead of 600; and Qwen with our server's speech detection relaxed (threshold 0.3, 2{,}000\,ms of silence, 300\,ms padding). The relaxed Qwen setting had no measurable cost: on the 46 quiet and 14 fan-on messages that completed (the longest timed out on our single server), WER was 10.3\% against 11.3\% with the default, and 5.6\% against 6.3\%, and it wrote no text on the 10 no-speech clips. Results are in Section~\ref{sec:whisper}.

\paragraph{Streaming and stability.} Audio is sent at the speed of speech in 32\,ms chunks at 16\,kHz (24\,kHz for Meta, GPT-Live and RT Whisper; 100\,ms chunks for AAI, which rejected chunks under 50\,ms, and Zoom). At the end the client signals end of audio and waits up to 30 seconds for the final text. Each clip uses a new session, from one client machine, with our own WebSocket code rather than vendor SDKs. Repeatability was checked on 42 clips played a week apart: gated Whisper and Gemini returned the same text, and Meta's WER was 16.5\% and 16.7\% (on the 41 clips it answered). Tables~\ref{tab:systems} and~\ref{tab:settings} give each system's model and settings; anything not listed is the provider's default.

\begin{table}[!htbp]
\centering\small
\begin{adjustbox}{max width=\textwidth}
\begin{tabular}{@{}llll@{}}
\toprule
System & Model & Interface & Audio sent \\
\midrule
Whisper SH & Whisper Large V3, our own deployment & WebSocket, client cuts segments & 16\,kHz, 32\,ms \\
Qwen & Qwen3-ASR 1.7B, our own deployment (one H100 MIG slice) & WebSocket & 16\,kHz, 32\,ms \\
Meta & \texttt{muse-voice-transcribe-1.0} & WebSocket & 24\,kHz, 32\,ms \\
Gemini, Gemini Smart & \texttt{gemini-3.5-transcribe-live} & Live API & 16\,kHz, 32\,ms \\
Deepgram & \texttt{nova-3} & WebSocket & 16\,kHz, 32\,ms \\
AAI 3.6, AAI 3.5 & \texttt{universal-3-6-pro}, \texttt{universal-3-5-pro} & WebSocket & 16\,kHz, 100\,ms \\
Zoom & Zoom Scribe (Live API) & WebSocket & 16\,kHz, 100\,ms \\
ElevenLabs, ElevenLabs NV & \texttt{scribe\_v2\_realtime} & WebSocket & 16\,kHz, 32\,ms \\
GPT-Live & \texttt{gpt-live-transcribe}, \texttt{delay}: \texttt{low} & Realtime API & 24\,kHz, 32\,ms \\
RT Whisper & \texttt{gpt-realtime-whisper} & Realtime API & 24\,kHz, 32\,ms \\
Gated Whisper & Whisper Large V3, our deployment with a client VAD gate & WebSocket, server cuts segments & 16\,kHz, 32\,ms \\
\bottomrule
\end{tabular}
\end{adjustbox}
\caption{Model, interface and audio format of each system. Our own deployments run on Baseten; their deployment identifiers are not given. Tests ran from one client machine.}
\label{tab:systems}
\end{table}

\begin{table}[!htbp]
\centering\scriptsize
\begin{adjustbox}{max width=\textwidth}
\begin{tabular}{@{}>{\raggedright\arraybackslash}p{2.0cm}>{\raggedright\arraybackslash}p{3.3cm}>{\raggedright\arraybackslash}p{3.4cm}>{\raggedright\arraybackslash}p{4.3cm}>{\raggedright\arraybackslash}p{3.4cm}@{}}
\toprule
System & Language setting & End of audio & When text is finalised & Other settings \\
\midrule
Whisper SH & \texttt{audio\_language}; two languages as \texttt{language\_\allowbreak options} & the client cuts the last segment & each segment the client cuts (Section~\ref{sec:whisperguide}) & beam width 5; partial text off \\
Gated Whisper & \texttt{audio\_language} & \texttt{end\_audio} & server's own speech detection (threshold 0.7, 480\,ms silence) & beam width 5; partial text off; client VAD in front \\
Qwen & \texttt{audio\_language} & \texttt{input\_audio\_buffer\allowbreak.commit} & server & partial text off \\
Meta & \texttt{languageBias} & \texttt{endStream} & once, at the end & push-to-talk mode; cumulative partial text \\
Gemini, Gemini Smart & \texttt{languageCodes} & \texttt{activityEnd} & once, at the end (automatic activity detection off) & \texttt{mode}: \texttt{VERBATIM} or \texttt{SMART} \\
Deepgram & \texttt{language}; \texttt{multi} for two languages or none & \texttt{CloseStream} & server's endpointing, at its default & \texttt{punctuate}, \texttt{smart\_format} \\
AAI 3.6, AAI 3.5 & \texttt{language\_codes} & \texttt{Terminate} & server's turn detection, at its default & \texttt{format\_turns} \\
Zoom & one locale per session & \texttt{session.close} & server's turn detection: 600\,ms of silence, at most 30\,s & partial text off; text clean-up off \\
ElevenLabs, ElevenLabs NV & \texttt{language\_code}; second language as \texttt{secondary\_\allowbreak languages} & a final empty chunk with \texttt{commit} & once, at the end (\texttt{commit\_strategy}: \texttt{manual}), though the server finalised the first 36\,s of longer messages itself (Table~\ref{tab:timing}); the \emph{pause} runs: at each 300\,ms pause (\texttt{vad}) & NV: \texttt{no\_verbatim} \\
GPT-Live & \texttt{languages} & \texttt{input\_audio\_buffer\allowbreak.commit} & once, at the end (server turn detection off) & \texttt{delay}: \texttt{low} \\
RT Whisper & \texttt{language} (one only) & \texttt{input\_audio\_buffer\allowbreak.commit} & once, at the end (server turn detection off) & -- \\
\bottomrule
\end{tabular}
\end{adjustbox}
\caption{Settings sent to each system. Parameter names are the providers' own.}
\label{tab:settings}
\end{table}

\begin{table}[!htbp]
\centering\small
\begin{tabular}{@{}lrrl@{}}
\toprule
& \texttt{low} & \texttt{minimal} & Difference [95\% CI] \\
\midrule
\multicolumn{4}{@{}l}{\emph{Our sets, WER}} \\
Quiet & 7.4 & 9.3 & $+$1.8 [1.3, 2.3] \\
Fan on & 3.8 & 5.6 & $+$1.8 [0.3, 3.4] \\
Whispered & 8.8 & 13.6 & $+$4.9 [2.4, 7.1] \\
Arabic MSA (Egypt) & 7.1 & 8.2 & $+$1.0 [$-$1.5, 3.4] \\
Saudi Arabic & 27.3 & 32.8 & $+$5.5 [1.4, 10.6] \\
Emirati Arabic & 53.5 & 57.1 & $+$3.5 [$-$1.9, 8.8] \\
German & 6.5 & 7.7 & $+$1.2 [$-$2.8, 4.9] \\
Spanish (Latin America) & 6.5 & 8.7 & $+$2.2 [$-$0.9, 5.9] \\
Spanish (Spain) & 8.4 & 12.1 & $+$3.7 [0.5, 7.6] \\
Spanish (Puerto Rico) & 3.1 & 6.8 & $+$3.7 [0.0, 9.2] \\
French & 9.4 & 14.5 & $+$5.1 [1.5, 8.6] \\
Hindi & 14.0 & 20.6 & $+$6.6 [3.1, 10.2] \\
Urdu & 21.5 & 19.7 & $-$1.8 [$-$14.3, 9.0] \\
Mandarin & 7.6 & 7.1 & $-$0.5 [$-$2.8, 1.6] \\[3pt]
\multicolumn{4}{@{}l}{\emph{EARS and VOiCES, WER}} \\
EARS, normal & 1.3 & 2.0 & $+$0.7 [0.5, 0.9] \\
EARS, whispered & 2.1 & 4.0 & $+$1.9 [1.5, 2.3] \\
VOiCES, clean & 3.5 & 4.7 & $+$1.3 [0.8, 1.8] \\
VOiCES, far & 8.0 & 12.3 & $+$4.3 [3.2, 5.4] \\
VOiCES, far + babble & 84.5 & 85.8 & $+$1.4 [0.2, 2.8] \\
VOiCES, far + TV & 29.4 & 35.8 & $+$6.4 [4.7, 8.4] \\[3pt]
\multicolumn{4}{@{}l}{\emph{Two languages, share of recordings with a language lost}} \\
EN\,+\,X, both selected & 6.0 & 6.0 & $+$0.0 [$-$3.0, 3.0] \\
EN\,+\,X, English only & 7.0 & 7.0 & $+$0.0 [$-$3.0, 3.0] \\
EN\,+\,X, auto-detect & 7.0 & 7.0 & $+$0.0 [$-$3.0, 3.0] \\
X\,+\,Y, both selected & 4.3 & 5.0 & $+$0.7 [$-$0.7, 2.0] \\
X\,+\,Y, auto-detect & 5.0 & 8.3 & $+$3.3 [1.3, 5.3] \\
\bottomrule
\end{tabular}
\caption{GPT-Live with its \texttt{delay} set to \texttt{low} (used in every other table) and to \texttt{minimal}, run together on each test set, so values can differ slightly from the main tables. WER (\%), scored as in Table~\ref{tab:overall} for our sets and with Whisper's English normaliser for EARS and VOiCES (Appendix~\ref{app:ext}); two-language rows are the share (\%) of all recordings in which a language went missing. Difference: \texttt{minimal} minus \texttt{low}, from unrounded values, with a paired bootstrap interval over clips (our sets), speakers (EARS and VOiCES) or recordings (two languages).}
\label{tab:gptdelay}
\end{table}

\section{Data and Scoring Details}\label{app:data}
\paragraph{The spoken reference.} The eleven systems' verbatim transcripts were aligned with the script, and an edit was proposed wherever enough systems heard the same departure: an added word when at least three did, a missing or changed word when at least five did (higher, because four systems drop the same whispered words). Proposals with at least four or six votes were pre-ticked. One team member listened to every message, accepted or rejected each proposal, and corrected six messages by hand. Complete words were kept as spoken, including self-corrected ones; cut-off fragments and fillers were left out, as in the scripts.

\paragraph{Correcting for many comparisons.} A difference is clear only if its paired 95\% interval excludes zero and its two-sided bootstrap p-value stays below 0.05 after a Holm correction within its group. The 17 groups are listed in \texttt{paper/analysis/stats\_review.ts} (4{,}000 resamples) and \texttt{gptdelay.py} (2{,}000): each English condition's best against next-best system; the whispered missing-word shares; the name gaps; script against spoken reference; the clean-up settings; ElevenLabs NV on Saudi and Emirati Arabic; English-only against both languages; ElevenLabs finalising; the Whisper setups (in-house and EARS); the EARS and VOiCES condition changes; and GPT-Live's delay settings. Two-language tests resample source clips (Section~\ref{sec:cs}). Differences outside these groups, such as the language averages, are not called clear.

\paragraph{Normalisation.} Capital letters and punctuation are ignored. For Arabic and Urdu, short-vowel marks and the elongation character are removed; for Arabic, the following are also merged: \textit{alif} with and without \textit{hamza}, \textit{ta marbuta} with \textit{ha}, and \textit{alif maqsura} with \textit{ya}. Traditional and simplified Chinese characters count as equal.

\paragraph{Mandarin.} With no spaces, each Chinese character and each run of Latin letters or digits counts as one unit: the mixed error rate (MER) of Mandarin--English code-switching work \citep{lyu2010seame}, not the match error rate of \citet{morris2004wer}. Chinese numerals in place of digits count as errors.

\paragraph{Language sets.} Most are read sentences (FLEURS and Puerto Rican Spanish); the Saudi and Emirati sets are dialect speech, and Spanish (Spain) is European Parliament speech.

\paragraph{Labelling names.} Each of the 184 different capitalised words was labelled by a written rule: names of people, organisations, products, services, programming languages, places and titled works count; days, months, titles, job titles, acronyms, headings and ordinary capitalised words do not. Only 156 of the 327 capitalised words are names. Adding 22 occurrences of listed names that the capital-letter rule misses (sentence starts, after ``Dr.'') and 5 lower-case uses of a listed product name gives 183.

\section{Accent Analysis}\label{app:accent}
Each recording was split into 4-second windows, which were labelled with the CommonAccent classifier \citep{zuluagagomez2023commonaccent}, keeping windows with a top-class probability of at least 0.7. On the 90 spoken messages, 62\% of windows are US English, 29\% South Asian English and 7\% British, New Zealand or Irish. Labels change between consecutive windows 28\% of the time, against 4\% on EARS, whose speakers each keep one accent; the reader confirmed the mix was deliberate. Whispered messages are 98\% US English, so the classifier hears no accent in whispers. Messages labelled mostly South Asian (25 of 90) have about the same WER as the rest for every system, but 16 of them are fan-on messages, so accent and condition cannot be separated.

\section{Additional Results}\label{app:extra}
\paragraph{Results by language.} Table~\ref{tab:overalllang} gives WER per language set. Without Urdu, Gemini has the lowest average (11.8\%); its paired intervals against every system except Gemini Smart (12.4\%) and AAI 3.6 (13.3\%) exclude zero, uncorrected for the 12 comparisons. With Urdu, AAI 3.6, RT Whisper, Gemini Smart, GPT-Live and Gemini (13.5--15.4\%) are within chance of each other.

\begin{table}[!htbp]
\centering\small
\begin{adjustbox}{max width=\textwidth}
\begin{tabular}{@{}lrrrrrrrrrrr|rr@{}}
\toprule
System & \shortstack[r]{Arabic MSA\\(Egypt)} & \shortstack[r]{Arabic\\Saudi} & \shortstack[r]{Arabic\\Emirati} & \shortstack[r]{\\German} & \shortstack[r]{Spanish\\Lat.\ Am.} & \shortstack[r]{Spanish\\Spain} & \shortstack[r]{Spanish\\P.\ Rico} & \shortstack[r]{\\French} & \shortstack[r]{\\Hindi} & \shortstack[r]{Mandarin\\(MER)} & \shortstack[r]{\\Urdu} & \shortstack[r]{Avg.\\no Urdu} & \shortstack[r]{Avg.\\with Urdu} \\
\midrule
Whisper SH & 9.2 & 37.1 & 60.0 & 6.9 & 5.7 & 19.5 & 5.8 & 6.5 & 29.8 & 4.1 & 19.7 & 18.4 & 18.6 \\
Qwen & 7.7 & 39.1 & 61.2 & 4.4 & 7.4 & 8.9 & 8.4 & 7.2 & 18.4 & 3.5 & 94.7 & 16.6 & 23.7 \\
Meta & 6.6 & 27.7 & 49.4 & 6.5 & 5.2 & 10.5 & 5.2 & 8.7 & 14.9 & 6.5 & 87.3$^*$ & 14.1 & 20.8 \\
Gemini & 2.6 & 29.3 & 48.8 & 5.2 & 6.5 & 8.4 & 3.7 & 7.2 & 3.1 & 3.3 & 51.8$^*$ & 11.8 & 15.4 \\
Gemini Smart & 3.6 & 35.2 & 49.4 & 4.8 & 5.7 & 7.4 & 4.7 & 6.9 & 2.6 & 3.8 & 32.4$^*$ & 12.4 & 14.2 \\
Deepgram & 11.2 & 38.3 & 57.1 & 8.1 & 8.7 & 9.5 & 4.2 & 10.5 & 20.6 & 3.3 & 14.4 & 17.1 & 16.9 \\
AAI 3.6 & 9.7 & 33.6 & 53.5 & 4.8 & 2.2 & 9.5 & 5.2 & 2.9 & 9.2 & 2.7 & 15.1 & 13.3 & 13.5 \\
AAI 3.5 & 14.3 & 34.8 & 57.1 & 6.0 & 5.2 & 10.0 & 5.2 & 4.0 & 25.9 & 3.5 & 91.5$^*$ & 16.6 & 23.4 \\
Zoom & 5.1 & 29.7 & 51.8 & 4.8 & 6.1 & 10.0 & 4.2 & 5.1 & 34.6 & 3.3 & 102.5$^*$ & 15.5 & 23.4 \\
ElevenLabs & 10.2 & 34.8 & 61.8 & 4.0 & 2.6 & 8.9 & 8.9 & 9.1 & 11.8 & 5.2 & 22.2 & 15.7 & 16.3 \\
ElevenLabs NV & 9.2 & 66.0 & 79.4 & 3.6 & 1.7 & 7.9 & 8.4 & 9.4 & 10.5 & 4.4 & 19.4 & 20.1 & 20.0 \\
GPT-Live & 6.6 & 27.7 & 51.8 & 4.4 & 6.5 & 8.9 & 5.8 & 9.4 & 11.8 & 6.5 & 19.7 & 14.0 & 14.5 \\
RT Whisper & 7.1 & 28.9 & 55.9 & 3.2 & 3.9 & 7.9 & 7.3 & 7.2 & 11.8 & 6.8 & 12.3 & 14.0 & 13.9 \\
\bottomrule
\end{tabular}
\end{adjustbox}
\caption{WER (\%) for each language set (20 clips each), lower is better. Nothing is in bold: in a paired comparison, the lowest WER in each set is never clearly below the next one. Mandarin uses the mixed error rate. Avg.: unweighted mean of the per-set WER over the 10 sets other than Urdu, and over all 11 with Urdu; the second counts each system's Urdu output as returned, including systems with no Urdu option, because that is what a user choosing Urdu receives. $^*$The system has no option for this language and detects it itself; not ranked.}
\label{tab:overalllang}
\end{table}

\paragraph{Failure of whispered dictations.} Table~\ref{tab:fwhisper} shows how many whispered clips failed and where the missing stretches fall.

\begin{table}[!htbp]
\centering\small
\begin{tabular}{@{}lrrrrr@{}}
\toprule
& & \multicolumn{2}{c}{Clips with WER} & Missing stretch & Cut-off \\
\cmidrule(lr){3-4}
& WER & $\geq$50\% & $\geq$90\% & at end & ending \\
\midrule
Whisper SH & 10.1 & 0 & 0 & 0 & 0 \\
Qwen & 46.6 & 38 & 25 & 6 & 12 \\
Meta & 11.3 & 0 & 0 & 0 & 0 \\
Gemini & 5.0 & 0 & 0 & 0 & 0 \\
Gemini Smart & 4.2 & 0 & 0 & 0 & 0 \\
Deepgram & 19.8 & 0 & 0 & 6 & 0 \\
AAI 3.6 & 30.5 & 31 & 19 & 12 & 25 \\
AAI 3.5 & 30.9 & 31 & 19 & 12 & 25 \\
Zoom & 34.7 & 31 & 12 & 6 & 0 \\
ElevenLabs & 9.1 & 0 & 0 & 0 & 0 \\
ElevenLabs NV & 7.1 & 0 & 0 & 6 & 0 \\
GPT-Live & 8.8 & 0 & 0 & 0 & 0 \\
RT Whisper & 8.9 & 0 & 0 & 0 & 0 \\
\bottomrule
\end{tabular}
\caption{How whispered dictations fail (16 clips); all values are percentages. Figure~\ref{fig:whispermix} splits the WER into wrong, missing and extra words. Clips with WER: share of clips with at least that WER. Missing stretch at end: share of clips with 5 or more words in a row missing in the last third. Cut-off ending: share of clips whose last 3 or more words are missing.}
\label{tab:fwhisper}
\end{table}

\paragraph{Normalised scores.} Table~\ref{tab:normwer} gives the re-scored English WER of Section~\ref{sec:numbers}. Quiet-English WER falls most for systems that write digits as words or keep fillers (Qwen 3.8 points, ElevenLabs 3.5, Deepgram 2.5), and Gemini Smart stays lowest.

\begin{table}[!htbp]
\centering\small
\begin{tabular}{@{}lrrr@{}}
\toprule
System & Quiet & Fan on & Whispered \\
\midrule
Whisper SH & 7.3 & 5.6 & 7.7 \\
Qwen & 10.2 & 4.6 & 43.6 \\
Meta & 7.4 & 4.3 & 7.6 \\
Gemini & 6.3 & 1.8 & 3.3 \\
Gemini Smart & 3.4 & 3.3 & 3.5 \\
Deepgram & 11.2 & 9.7 & 16.4 \\
AAI 3.6 & 6.8 & 3.5 & 29.4 \\
AAI 3.5 & 7.4 & 3.8 & 30.1 \\
Zoom & 6.0 & 2.6 & 33.8 \\
ElevenLabs & 9.8 & 4.4 & 7.1 \\
ElevenLabs NV & 4.3 & 3.3 & 5.6 \\
GPT-Live & 5.7 & 2.1 & 6.7 \\
RT Whisper & 7.0 & 3.6 & 6.7 \\
\bottomrule
\end{tabular}
\caption{English WER (\%) after Whisper's English text normaliser \citep{radford2023whisper}, applied to reference and transcript alike. Tokenisation differs slightly from Table~\ref{tab:overall}.}
\label{tab:normwer}
\end{table}

\paragraph{Disfluency-removal settings.} Beyond quiet English (Section~\ref{sec:cleanup}), Gemini and Gemini Smart are close with the fan on and whispered, and Smart is worse when punctuation and capitals are scored. Smart lowers Urdu from 51.8\% to 32.4\% (Section~\ref{sec:script}) and raises Saudi Arabic from 29.3\% to 35.2\%. ElevenLabs NV also helps with the fan on and whispered, but it clearly worsens Saudi Arabic (34.8\% to 66.0\%; difference 31.3 points, 95\% CI 21.5 to 40.2) and Emirati Arabic (61.8\% to 79.4\%; 17.6 points, 95\% CI 9.1 to 25.9).

\paragraph{Ranks against the two references.} Across the thirteen rows of Table~\ref{tab:verbatim}, the rank correlation between script and spoken reference is 0.93 on quiet English, 0.78 with the fan on and 0.98 whispered.

\paragraph{Failed clips by language.} Table~\ref{tab:fworst} counts each system's failed clips in each set.

\begin{table}[!htbp]
\centering\small
\begin{adjustbox}{max width=\textwidth}
\begin{tabular}{@{}lrrrrrr@{}}
\toprule
System & Saudi & Emirati & Hindi & Urdu & Other & Total \\
\midrule
Whisper SH & 25 & 60 & 20 & 0 & 2 & 11 \\
Qwen & 30 & 75 & 10 & 95 & 1 & 20 \\
Meta & 20 & 50 & 15 & 85 & 1 & 16 \\
Gemini & 20 & 55 & 0 & 45 & 1 & 12 \\
Gemini Smart & 20 & 55 & 0 & 25 & 1 & 10 \\
Deepgram & 30 & 70 & 10 & 0 & 1 & 10 \\
AAI 3.6 & 30 & 60 & 5 & 0 & 1 & 10 \\
AAI 3.5 & 35 & 70 & 25 & 90 & 1 & 21 \\
Zoom & 15 & 55 & 20 & 100 & 1 & 18 \\
ElevenLabs & 25 & 70 & 10 & 5 & 1 & 10 \\
ElevenLabs NV & 75 & 80 & 5 & 0 & 1 & 15 \\
GPT-Live & 15 & 55 & 5 & 5 & 1 & 8 \\
RT Whisper & 20 & 55 & 10 & 0 & 2 & 9 \\
\bottomrule
\end{tabular}
\end{adjustbox}
\caption{Share of non-English clips (\%) with a WER of 50\% or more, by set (20 clips per set). Other: the other seven sets together. Total: all 220 non-English clips.}
\label{tab:fworst}
\end{table}

\paragraph{Clips hard for one system only.} Of 326 speech clips, 23 are above 30\% for exactly one system: 9 for Qwen, 6 Deepgram, 3 each Zoom and ElevenLabs, and 2 Whisper SH (stock phrases on Spanish (Spain); Section~\ref{sec:whisperguide}).

\paragraph{Arabic character error rates.} CER is far below WER on Saudi (11--27\%) and Emirati Arabic (15--27\%), consistent with spelling variation but not proof of it, since attached prefixes keep CER far below WER on Arabic MSA (Egypt) too (0.5--4.3\% against 2.6--14.3\%).

\paragraph{Hindi.} AAI 3.5 fails on 25\% of Hindi clips, Whisper SH and Zoom on 20\%, Gemini on none (lowest WER: 3.1\%, 2.6\% with Smart). As for Zoom (Section~\ref{sec:script}), Whisper SH's Hindi CER is less than half its WER (13.6\% against 29.8\%), and almost so for AAI 3.5. Merging common Devanagari spelling variants lowers any system's Hindi WER by at most 2.2 points.

\paragraph{Types of substitution.} Table~\ref{tab:fsubkind} sorts wrong English words into numbers written the other way, names, near-misses (Table~\ref{tab:taxonomy}) and other words.

\begin{table}[!htbp]
\centering\small
\begin{adjustbox}{max width=\textwidth}
\begin{tabular}{@{}lrrrrr@{}}
\toprule
& Wrong words & \multicolumn{4}{c}{Share of wrong words} \\
\cmidrule(l){3-6}
System & (\% of words) & Number & Name & Near-miss & Other word \\
\midrule
Whisper SH & 3.6 & 14 & 5 & 23 & 58 \\
Qwen & 5.7 & 31 & 4 & 14 & 52 \\
Meta & 4.0 & 30 & 4 & 19 & 46 \\
Gemini & 2.1 & 24 & 5 & 22 & 48 \\
Gemini Smart & 1.8 & 30 & 4 & 19 & 47 \\
Deepgram & 6.6 & 15 & 4 & 18 & 63 \\
AAI 3.6 & 2.8 & 27 & 3 & 23 & 47 \\
AAI 3.5 & 2.9 & 26 & 4 & 19 & 52 \\
Zoom & 2.8 & 18 & 4 & 21 & 57 \\
ElevenLabs & 3.0 & 18 & 4 & 22 & 56 \\
ElevenLabs NV & 2.7 & 22 & 3 & 22 & 52 \\
GPT-Live & 3.3 & 35 & 5 & 14 & 46 \\
RT Whisper & 3.3 & 25 & 5 & 20 & 50 \\
\bottomrule
\end{tabular}
\end{adjustbox}
\caption{What the wrong words are, in all English dictation (quiet, fan on and whispered together). Wrong words: share of all words. The other columns: share (\%) of the system's wrong words in each group. Number: a number written in digits where the script has words, or the other way round. Near-miss: as in Table~\ref{tab:taxonomy}.}
\label{tab:fsubkind}
\end{table}

\paragraph{Number habits.} Table~\ref{tab:fnumbers} shows how often each system keeps or switches each number form.

\begin{table}[!htbp]
\centering\small
\begin{adjustbox}{max width=\textwidth}
\begin{tabular}{@{}lrrrrr@{}}
\toprule
& \multicolumn{2}{c}{Digits in script} & \multicolumn{2}{c}{Words in script} & Share of \\
\cmidrule(lr){2-3}\cmidrule(lr){4-5}
System & kept & as words & kept & as digits & errors \\
\midrule
Whisper SH & 79 & 15 & 81 & 7 & 6.1 \\
Qwen & 13 & 68 & 93 & 0 & 10.4 \\
Meta & 48 & 47 & 93 & 1 & 12.9 \\
Gemini & 79 & 19 & 89 & 2 & 8.4 \\
Gemini Smart & 86 & 12 & 74 & 12 & 14.0 \\
Deepgram & 47 & 37 & 87 & 4 & 7.2 \\
AAI 3.6 & 95 & 0 & 47 & 42 & 7.2 \\
AAI 3.5 & 95 & 0 & 48 & 41 & 6.7 \\
Zoom & 79 & 13 & 79 & 10 & 5.0 \\
ElevenLabs & 75 & 21 & 88 & 2 & 4.8 \\
ElevenLabs NV & 75 & 22 & 89 & 1 & 11.2 \\
GPT-Live & 50 & 45 & 95 & 0 & 16.1 \\
RT Whisper & 65 & 31 & 92 & 1 & 10.2 \\
\bottomrule
\end{tabular}
\end{adjustbox}
\caption{How each system writes numbers in English dictation (\%). Share of errors: number-format switches as a share of all the system's English errors. This is a minimum, because a number made of several words counts once. Kept and switched do not add up to 100\%: the remainder were misheard or missing.}
\label{tab:fnumbers}
\end{table}

\subsection{Latency}\label{app:speed}
Table~\ref{tab:latency} gives two delays from the test client. The \emph{wait} runs from the client's end-of-audio signal to the service's last message, so it excludes endpointing. That last message is the final text for Meta, Gemini, ElevenLabs, GPT-Live and RT Whisper, the end-of-session message for AAI and Zoom, the closed connection for Deepgram, and an end-of-audio message for our own deployments; for the session-message cases the wait can run slightly past the last text. The \emph{connect} time is how long a session takes to open. The median (p50) and a slow case (p90) are given, with up to six sessions at a time per system. The AAI models have the slowest p90 (over 1{,}200\,ms). Only Gemini's wait grows with clip length (591\,ms under 20 seconds, 1{,}084\,ms over a minute). Repeat runs moved waits by up to 320\,ms, so treat them as indicative.

\begin{table}[!htbp]
\centering\small
\begin{adjustbox}{max width=\textwidth}
\begin{tabular}{@{}lrrrr@{}}
\toprule
& \multicolumn{2}{c}{Wait (ms)} & \multicolumn{2}{c}{Connect (ms)} \\
\cmidrule(lr){2-3}\cmidrule(l){4-5}
System & p50 & p90 & p50 & p90 \\
\midrule
Whisper SH & 361 & 498 & 837 & 940 \\
Qwen & 388 & 542 & 1592 & 4428 \\
Meta & 313 & 371 & 1368 & 1756 \\
Gemini & 584 & 905 & 674 & 814 \\
Gemini Smart & 434 & 703 & 718 & 1201 \\
Deepgram & 357 & 431 & 911 & 1053 \\
AAI 3.6 & 722 & 1210 & 1227 & 1399 \\
AAI 3.5 & 727 & 1218 & 1228 & 1370 \\
Zoom & 407 & 616 & 1406 & 1655 \\
ElevenLabs & 398 & 493 & 685 & 765 \\
ElevenLabs NV & 395 & 459 & 660 & 727 \\
GPT-Live & 760 & 858 & 1207 & 1374 \\
RT Whisper & 735 & 832 & 1185 & 1312 \\
\bottomrule
\end{tabular}
\end{adjustbox}
\caption{Delays over all single-language sessions. Wait: from the end of the audio to the final text. Connect: from opening the connection to the session being ready. p50: median; p90: 90\% of sessions were faster.}
\label{tab:latency}
\end{table}

\paragraph{Finalisation timing and network round trip.} A timing run on the 106 English messages recorded when text was first finalised and the round trip to each provider (Table~\ref{tab:timing}). Whisper SH, Qwen, Deepgram, Zoom and AAI finalise text during speech on 83--99\% of messages, typically 5--9 seconds in; Meta, Gemini, GPT-Live and RT Whisper finalise only at the end, as set up. ElevenLabs finalised the first 36 seconds early on every message longer than about 35 seconds. Subtracting the round trip moves Deepgram from seventh to third, but the same six rows stay slowest, at 594--716\,ms beyond it.

\begin{table}[!htbp]
\centering\small
\begin{adjustbox}{max width=\textwidth}
\begin{tabular}{@{}lrrrrr@{}}
\toprule
& Final text & First final & Wait & Round & Wait minus \\
System & during speech (\%) & text (s) & p50 (ms) & trip (ms) & round trip (ms) \\
\midrule
Whisper SH & 95 & 8.5 & 400 & 276 & 124 \\
Qwen & 96 & 7.0 & 360 & 276 & 84 \\
Meta & 0 & -- & 433 & 56 & 377 \\
Gemini & 0 & -- & 739 & 58 & 681 \\
Gemini Smart & 0 & -- & 754 & 58 & 696 \\
Deepgram & 99 & 5.3 & 522 & 309 & 213 \\
AAI 3.6 & 83 & 8.7 & 841 & 200 & 641 \\
AAI 3.5 & 85 & 8.6 & 794 & 200 & 594 \\
Zoom & 84 & 8.8 & 429 & 28 & 401 \\
ElevenLabs & 53 & 35.9 & 392 & 145 & 246 \\
ElevenLabs NV & 52 & 35.9 & 376 & 145 & 230 \\
GPT-Live & 0 & -- & 728 & 28 & 700 \\
RT Whisper & 0 & -- & 744 & 28 & 716 \\
\bottomrule
\end{tabular}
\end{adjustbox}
\caption{Timing run on the 106 English messages. Final text during speech: share of messages with finalised text before the audio ended. First final text: median time from the start of the audio to the first finalised text, over those messages. Round trip: median TCP handshake time to the provider during the runs, about 300 samples each; for services behind a content-delivery network (OpenAI, Zoom, and in part Gemini and Meta) it reaches only the nearest edge server, so it understates the round trip to the model.}
\label{tab:timing}
\end{table}

\section{External Corpora}\label{app:ext}
\paragraph{The EARS speakers.} The 107 speakers are 60 women, 43 men and 4 other or not stated, aged 18 to 75, mostly native speakers of American English.

\paragraph{The VOiCES clips.} One hundred sentences are taken, one per speaker (50 women, 50 men), as the clean original and as recorded by a far studio microphone in room 4 in quiet, with babble and with a television: 400 clips, 26 minutes per version.

\paragraph{Scoring.} VOiCES references are LibriSpeech text (lower case, numbers as words), so for both corpora Whisper's English normaliser is applied to reference and transcript and only WER is reported, with intervals over speakers. Clips gated Whisper's VAD never sent, and sessions with no transcript within 30 seconds (only Gemini, all on noisy clips), count as empty. One traced babble session stayed open and returned ``Add, and,'' 17 seconds after the audio ended; the others were not traced.

\paragraph{Far-field and noise results.} Clean recordings score 2.1--4.6\%. A far microphone in quiet adds under 3 points for Whisper SH, Gemini, the AAI models and ElevenLabs, 4--6 for Meta, Deepgram, GPT-Live, RT Whisper and Zoom, and 8--10 for Qwen and gated Whisper. Babble adds 31--85 points, mostly as missing words for eleven of the fourteen rows, while extra words stay under 7.3\%, so no system writes down the background talk. Gemini Smart also gave no text on 12 babble clips; leaving those out, its babble WER is 67.1\% (Gemini's: Section~\ref{sec:voices}), a lower bound on easier clips. A television raises WER less and unevenly: Zoom by 2.6 points (within chance), the AAI models by 7.4--8.3, RT Whisper by 32 and Deepgram by 40.

\begin{table}[!htbp]
\centering\small
\begin{adjustbox}{max width=\textwidth}
\begin{tabular}{@{}lrrrrrr@{}}
\toprule
& & Far & Far + & Far + & Far + babble: & Babble: \\
System & Clean & mic & babble & TV & wrong / missing / extra & no text \\
\midrule
Whisper SH & 2.1 & 4.2 & 62.5 & 17.7 & 17.7 / 41.6 / 3.3 & 0 \\
Gated Whisper & 2.1 & 11.9 & 63.4 & 23.4 & 24.7 / 34.2 / 4.5 & 1 \\
Qwen & 2.3 & 10.5 & 50.9 & 16.3 & 27.7 / 21.9 / 1.3 & 0 \\
Meta & 2.4 & 6.5 & 77.5 & 25.6 & 6.7 / 70.4 / 0.3 & 0 \\
Gemini & 2.5 & 5.1 & 81.7 & 24.8 & 4.4 / 76.3 / 1.0 & 67 \\
Gemini Smart & 4.6 & 5.3 & 71.2 & 23.0 & 22.4 / 45.4 / 3.4 & 12 \\
Deepgram & 4.3 & 10.1 & 94.9 & 50.4 & 1.3 / 93.5 / 0.1 & 0 \\
AAI 3.6 & 2.3 & 4.8 & 59.6 & 13.2 & 12.1 / 45.5 / 2.0 & 0 \\
AAI 3.5 & 2.2 & 5.0 & 52.5 & 12.4 & 20.9 / 27.2 / 4.3 & 0 \\
Zoom & 2.3 & 7.7 & 38.7 & 10.3 & 12.8 / 24.3 / 1.5 & 0 \\
ElevenLabs & 2.8 & 4.6 & 66.6 & 19.4 & 33.9 / 25.7 / 7.0 & 0 \\
ElevenLabs NV & 2.7 & 4.6 & 66.9 & 18.8 & 32.4 / 27.2 / 7.3 & 0 \\
GPT-Live & 3.2 & 8.0 & 85.1 & 29.0 & 6.6 / 78.0 / 0.5 & 0 \\
RT Whisper & 3.0 & 8.8 & 93.6 & 41.2 & 2.0 / 91.6 / 0.0 & 0 \\
\bottomrule
\end{tabular}
\end{adjustbox}
\caption{VOiCES: 100 sentences from 100 speakers, each as the clean original and as recorded by a far microphone in one room, with no background sound, with babble and with TV. WER (\%), numbers and spellings normalised. Babble, no text: of the 100 babble clips, those with no transcript within 30 seconds, counted as empty.}
\label{tab:voices}
\end{table}

\section{Code-Switching Analysis}\label{app:cs}
\paragraph{Screened recordings.} A part may be lost because it is hard on its own. Table~\ref{tab:csfair} repeats Table~\ref{tab:csoverall} on the 85 EN\,+\,X and 232 X\,+\,Y recordings whose parts every system gets at least 50\% right alone; since the screen uses the tested systems, it keeps easy recordings. Rates are the same or up to 10 points lower, except ElevenLabs NV with automatic detection on EN\,+\,X (79\% to 80\%). The order is the same except that Meta and AAI 3.6 swap on X\,+\,Y with both selected. Adding Qwen removed no recording.

\begin{table}[!htbp]
\centering\small
\begin{adjustbox}{max width=\textwidth}
\begin{tabular}{@{}lrrrrr@{}}
\toprule
& \multicolumn{3}{c}{EN\,+\,X (85 recordings)} & \multicolumn{2}{c}{X\,+\,Y (232 recordings)} \\
\cmidrule(lr){2-4}\cmidrule(l){5-6}
System & Both selected & English only & Auto-detect & Both selected & Auto-detect \\
\midrule
Whisper SH & 39 & 85 & 42 & 87 & 87 \\
Qwen & 14 & 33 & 14 & 43 & 43 \\
Meta & 13 & 16 & 15 & 22 & 25 \\
Gemini & 42 & 92 & 66 & 27 & 69 \\
Gemini Smart & 46 & 85 & 68 & 55 & 80 \\
Deepgram & 32 & 99 & 32 & 53 & 53 \\
AAI 3.6 & 4 & 19 & 8 & 24 & 21 \\
AAI 3.5 & 6 & 29 & 11 & 19 & 25 \\
Zoom & 56 & 56 & 56 & 81 & 81 \\
ElevenLabs & 61 & 84 & 61 & 92 & 91 \\
ElevenLabs NV & 80 & 93 & 80 & 96 & 96 \\
ElevenLabs, pause & \textbf{1} & 36 & \textbf{1} & 36 & 36 \\
ElevenLabs NV, pause & 2 & 46 & 2 & 62 & 63 \\
GPT-Live & \textbf{1} & \textbf{1} & \textbf{1} & \textbf{0} & \textbf{0} \\
RT Whisper & 5 & 5 & 4 & 4 & 5 \\
\bottomrule
\end{tabular}
\end{adjustbox}
\caption{As Table~\ref{tab:csoverall}, on the 85 EN\,+\,X and 232 X\,+\,Y recordings kept by the screen; lowest in each column in bold. The \emph{pause} rows are ElevenLabs and ElevenLabs NV set to finalise at each pause of 300\,ms, run on these recordings only.}
\label{tab:csfair}
\end{table}

\paragraph{ElevenLabs finalisation.} Finalising once, ElevenLabs loses the non-English part even of Spanish recordings it transcribes alone. Set to finalise at each 300\,ms pause on the screened recordings (the same gap placed between clips, so this favours it), recordings with a lost language fell from 61\% to 1\% with English (95\% CI 0 to 7\%) and from 92\% to 36\% without (95\% CI 26 to 47\%); for ElevenLabs NV, from 80\% to 2\% and from 96\% to 62\%. Without English, the part still lost is almost always the second (80 of 83).

\paragraph{Language requests.} With both languages selected, systems that accept several are given both. Deepgram uses its multi-language mode in both settings, which covers English, Spanish, French, German, Hindi, Russian, Portuguese, Japanese, Italian and Dutch \citep{deepgramdocs}, not Arabic or Mandarin. RT Whisper is given no language in either setting, and Zoom none (its default behaves like English), so for these three the two settings send the same request; results matched except for RT Whisper on 3 recordings. Qwen is sent both languages as \texttt{language\_options}, but wrote the same text in both settings on all 400 recordings.

\begin{table}[!htbp]
\centering\small
\begin{adjustbox}{max width=\textwidth}
\begin{tabular}{@{}llrrrrrrr@{}}
\toprule
Setting & System & Non-English missing & First missing & Second missing & Nothing & Other text & Romanised & Translated \\
\midrule
Both selected & Whisper SH & 33 & 25 & 20 & 43 & 2 & 0 & 0 \\
 & Qwen & 20 & 6 & 14 & 6 & 14 & 8 & 8 \\
 & Meta & 15 & 0 & 19 & 18 & 1 & 0 & 0 \\
 & Gemini & 38 & 35 & 16 & 45 & 6 & 10 & 0 \\
 & Gemini Smart & 46 & 23 & 29 & 49 & 3 & 6 & 0 \\
 & Deepgram & 40 & 17 & 23 & 18 & 22 & 34 & 4 \\
 & AAI 3.6 & 9 & 3 & 7 & 5 & 5 & 4 & 0 \\
 & AAI 3.5 & 12 & 5 & 8 & 7 & 6 & 2 & 2 \\
 & Zoom & 62 & 30 & 32 & 55 & 7 & 6 & 0 \\
 & ElevenLabs & 60 & 25 & 37 & 59 & 3 & 0 & 2 \\
 & ElevenLabs NV & 77 & 34 & 46 & 73 & 7 & 2 & 2 \\
 & GPT-Live & 6 & 0 & 6 & 4 & 2 & 2 & 0 \\
 & RT Whisper & 10 & 2 & 9 & 8 & 3 & 0 & 0 \\[3pt]
English only & Whisper SH & 86 & 44 & 43 & 64 & 23 & 4 & 18 \\
 & Qwen & 38 & 20 & 18 & 17 & 21 & 10 & 14 \\
 & Meta & 20 & 0 & 23 & 22 & 1 & 0 & 0 \\
 & Gemini & 90 & 50 & 43 & 92 & 1 & 0 & 0 \\
 & Gemini Smart & 86 & 36 & 51 & 85 & 2 & 2 & 2 \\
 & Deepgram & 99 & 50 & 49 & 94 & 5 & 4 & 0 \\
 & AAI 3.6 & 25 & 10 & 15 & 18 & 7 & 6 & 2 \\
 & AAI 3.5 & 35 & 13 & 22 & 31 & 4 & 4 & 0 \\
 & Zoom & 62 & 30 & 32 & 55 & 7 & 6 & 0 \\
 & ElevenLabs & 85 & 50 & 35 & 82 & 3 & 2 & 2 \\
 & ElevenLabs NV & 93 & 50 & 43 & 86 & 7 & 2 & 4 \\
 & GPT-Live & 7 & 0 & 7 & 5 & 2 & 2 & 0 \\
 & RT Whisper & 12 & 3 & 9 & 9 & 3 & 0 & 0 \\[3pt]
Auto-detect & Whisper SH & 37 & 27 & 22 & 43 & 6 & 2 & 4 \\
 & Qwen & 20 & 6 & 14 & 6 & 14 & 8 & 8 \\
 & Meta & 17 & 0 & 21 & 20 & 1 & 0 & 0 \\
 & Gemini & 36 & 54 & 19 & 70 & 3 & 4 & 0 \\
 & Gemini Smart & 59 & 28 & 45 & 72 & 1 & 0 & 0 \\
 & Deepgram & 40 & 17 & 23 & 18 & 22 & 34 & 4 \\
 & AAI 3.6 & 13 & 3 & 12 & 10 & 5 & 2 & 0 \\
 & AAI 3.5 & 13 & 5 & 11 & 11 & 5 & 0 & 2 \\
 & Zoom & 62 & 30 & 32 & 55 & 7 & 6 & 0 \\
 & ElevenLabs & 60 & 25 & 37 & 59 & 3 & 0 & 2 \\
 & ElevenLabs NV & 76 & 33 & 46 & 73 & 6 & 2 & 2 \\
 & GPT-Live & 7 & 1 & 6 & 6 & 1 & 0 & 0 \\
 & RT Whisper & 10 & 2 & 8 & 6 & 4 & 0 & 0 \\[3pt]
\bottomrule
\end{tabular}
\end{adjustbox}
\caption{Missing parts in EN\,+\,X recordings, as a share (\%) of all 100 recordings. Non-English missing: the part in the other language is missing (the measure of the filter and preference rule in Section~\ref{sec:cs}). Nothing: no text where the part should be. Other text: different text in its place. Romanised and Translated are measured only on the 50 recordings whose other language has its own writing system (Arabic, Hindi or Mandarin), as a share of those 50. Romanised: the part is replaced by Latin-letter text that is not English. Translated: replaced by English words.}
\label{tab:fcs}
\end{table}

\section{Whisper Deployment Results}\label{app:whisper}
\paragraph{Cutting at pauses.} With its VAD off, the gated deployment drops 5 or more words from the last third of 44\% of whispered clips, and the self-hosted deployment has no cut-off endings; on EARS they miss 30.5\% and 10.4\% of whispered words (Table~\ref{tab:earsvad}). The gated server's own speech detection, which was not measured, may account for part of the difference. Cutting at pauses is as fast: 361 against 360\,ms typical, 498 against 596\,ms slow case.

\paragraph{Invented text.} The two stock phrases (Finding 4) give the self-hosted deployment 19.5\% WER on Spanish (Spain), twice the gated setup's 10.0\%. Not sending segments with no speech stops text on silence (Whisper-1 split wrote none), but also lost whispered clips the VAD missed (23.4\% against 7.7\% for Whisper-1).

\paragraph{The nine setups in full.}
\begin{enumerate}\itemsep1pt
  \item \textbf{Gated, VAD on (``gated Whisper'').} A deployment on Baseten with a voice gate on the client: a Silero VAD in front of Whisper drops audio it judges to be silence: it waits for speech above a 0.5 probability, keeps 96\,ms before it, sends at most 512\,ms of any pause, and stops sending after 544\,ms of silence. The deployment cuts the audio into segments itself, with its own server-side speech detection, which is not under our control.
  \item \textbf{Gated, VAD off.} The same deployment, sent all of the audio: the client's Silero VAD is off, but the server's own speech detection still runs.
  \item \textbf{Self-hosted, Silero cuts (``Whisper SH'').} A second deployment of our own, which does no segmentation. The client runs the same VAD only to decide where each segment ends: at a pause of 544\,ms after speech, once at least 5 seconds have been sent, and always before 30 seconds. It then tells the server to transcribe that segment. Every sample of audio is sent.
  \item \textbf{Self-hosted, Silero drops.} The same deployment, with Silero VAD also dropping audio as in setup 1.
  \item \textbf{Self-hosted, TEN cuts.} Setup 3 with TEN VAD \citep{tenvad}, an open VAD from the TEN framework, in place of Silero, at its default threshold of 0.5. Every sample is sent.
  \item \textbf{Self-hosted, TEN drops.} Setup 4 with TEN VAD: the same rules decide which audio is dropped, using TEN's speech probabilities instead of Silero's.
  \item \textbf{OpenAI GPT-Realtime-Whisper (``RT Whisper'').} OpenAI's hosted streaming Whisper.
  \item \textbf{OpenAI Whisper-1.} OpenAI's file transcription, so the whole recording is uploaded when the user stops.
  \item \textbf{Whisper-1 split.} Whisper-1 with the client cutting the recording at pauses, using the VAD rule of setup 3, and uploading each segment as soon as it is cut, with the previous segment's text as context. A segment in which the VAD heard no speech is not sent.
\end{enumerate}

\begin{table}[!htbp]
\centering\small
\begin{adjustbox}{max width=\textwidth}
\begin{tabular}{@{}llrrrr@{}}
\toprule
& Set & VAD on & VAD off & Change & 95\% CI \\
\midrule
Gated Whisper & Quiet & 8.7 & 9.5 & +0.9 & [+0.1, +1.6] \\
 & Fan on & 7.0 & 7.5 & +0.6 & [$-$1.3, +2.5] \\
 & Whispered & 80.7 & 61.1 & $-$19.6 & [$-$34.5, $-$7.5] \\
 & Arabic (Emirati) & 62.9 & 58.2 & $-$4.7 & [$-$10.6, $-$0.6] \\
 & Whispered clips never sent & 12.5 & 0.0 & & \\[3pt]
\bottomrule
\end{tabular}
\end{adjustbox}
\caption{Gated Whisper with the client's Silero VAD on and off (the server's own speech detection runs in both), WER (\%). Change: VAD off minus VAD on, in percentage points, with its 95\% confidence interval from a paired comparison. Whispered clips never sent: share of the 16 whispered clips the VAD never sent.}
\label{tab:vad}
\end{table}

\paragraph{The same setups across 107 speakers.} Table~\ref{tab:earsvad} repeats the dropping and cutting setups on EARS: all are at 1.0--1.8\% on normal speech and separate on whispered speech as on our clips.

\begin{table}[!htbp]
\centering\small
\begin{adjustbox}{max width=\textwidth}
\begin{tabular}{@{}lrrrrrr@{}}
\toprule
& & & Change & Whispered: & Clips & Speakers \\
Setup & Normal & Whispered & (paired) & missing & never sent & $\geq$20\% \\
\midrule
Gated, VAD on & 1.0 & 42.9 {\scriptsize [36.0, 49.7]} & +31.4 {\scriptsize [+25.3, +37.5]} & 39.1 & 38 & 67 \\
Gated, VAD off & 1.2 & 32.8 {\scriptsize [26.2, 39.1]} & +21.3 {\scriptsize [+15.6, +27.6]} & 30.5 & 0 & 53 \\
Self-hosted, Silero cuts & 1.5 & 11.5 {\scriptsize [8.2, 15.2]} & -- & 10.4 & 0 & 20 \\
Self-hosted, Silero drops & 1.5 & 38.3 {\scriptsize [31.8, 45.0]} & +26.8 {\scriptsize [+21.3, +32.2]} & 36.1 & 0 & 62 \\
Self-hosted, TEN cuts & 1.3 & 9.5 {\scriptsize [6.5, 13.1]} & $-$2.0 {\scriptsize [$-$3.7, $-$0.3]} & 8.6 & 0 & 17 \\
Self-hosted, TEN drops & 1.8 & 14.4 {\scriptsize [10.1, 19.0]} & +2.9 {\scriptsize [$-$0.4, +6.7]} & 13.3 & 0 & 24 \\
\bottomrule
\end{tabular}
\end{adjustbox}
\caption{Whisper setups on EARS: 107 speakers, 428 normal and 428 whispered clips. WER (\%), numbers and spellings normalised as in Table~\ref{tab:ears}, with 95\% intervals from resampling speakers. Change: whispered WER minus that of the self-hosted deployment with Silero only cutting, paired by clip. Whispered, missing: missing words as a share of all words. Clips never sent: whispered clips the VAD never sent. Speakers $\geq$20\%: speakers whose whispered WER is 20\% or more.}
\label{tab:earsvad}
\end{table}

\paragraph{TEN VAD, further results.} TEN does not keep whispered speech by calling everything speech: on fan noise, steady noise, mains hum and television it calls about as much audio speech as Silero, and at most 2.2\% of the windows of any no-speech clip. It does call much more background talk speech: with babble, 97\% of windows against 80\% for Silero, 15 points more than on the same recordings without babble; what Whisper writes when TEN passes that audio was not tested. On EARS, TEN also beats Silero when only cutting (9.5\% against 11.5\%, difference 95\% CI $-3.7$ to $-0.3$), because it calls more whispered audio speech; on no-speech clips both call almost none (0.4\% and 0\%). Dropping with TEN gives the self-hosted deployment 7.2\% on quiet English (8.3\% sending all audio, within its interval), 17.6\% averaged over other languages, 14.7\% on Spanish (Spain) instead of 19.5\%, and text on 2 no-speech clips instead of 6. The cost is a slightly longer wait (395 against 361\,ms) and more lost languages (54\% against 45\% with English, 96\% against 87\% without).

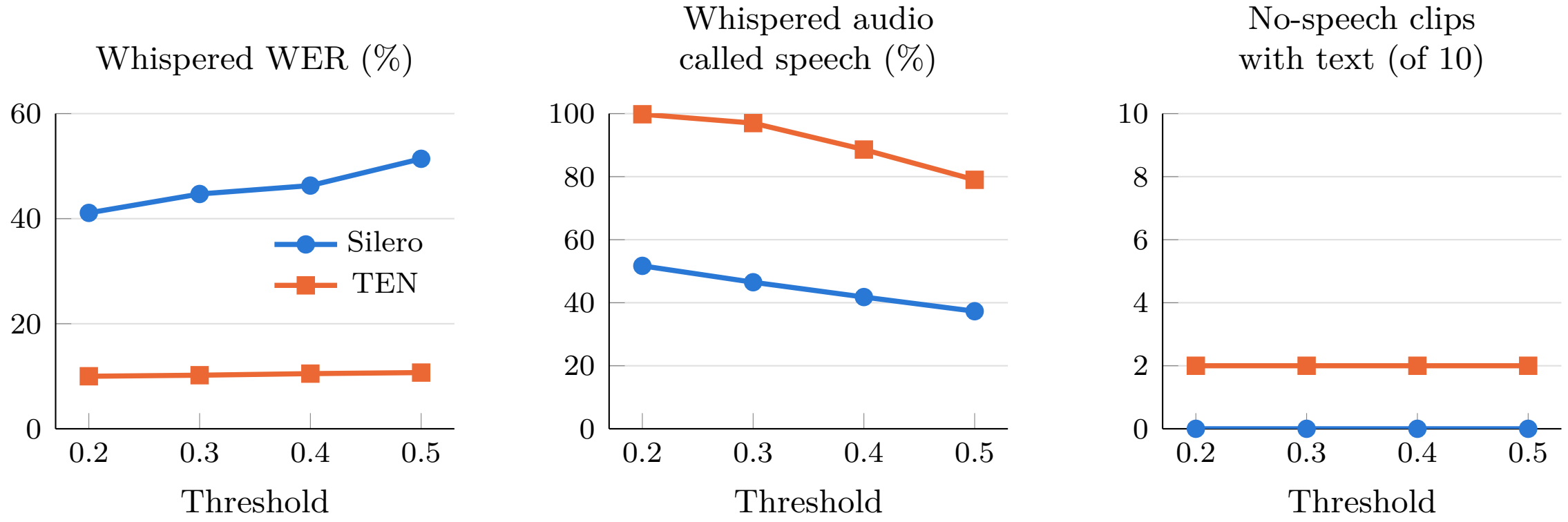
\begin{figure}[!htbp]
\centering
\begin{tikzpicture}
\begin{axis}[at={(0cm,0)}, width=5.4cm, height=4.6cm, title={Whispered WER (\%)}, title style={font=\footnotesize, align=center},
  xtick={0.2,0.3,0.4,0.5}, xlabel={Threshold}, xlabel style={font=\footnotesize}, ticklabel style={font=\scriptsize}, ymin=0, ymax=60,
  axis x line*=bottom, axis y line*=left, ymajorgrids, major grid style={black!12}, legend style={draw=none, fill=none, font=\scriptsize, at={(0.97,0.52)}, anchor=east}]
\addplot[figblue, line width=1.4pt, mark=*, mark size=1.8pt] coordinates {(0.2,41.1) (0.3,44.7) (0.4,46.3) (0.5,51.4)};
\addplot[figorange, line width=1.4pt, mark=square*, mark size=1.8pt] coordinates {(0.2,10.0) (0.3,10.2) (0.4,10.5) (0.5,10.7)};
\legend{Silero, TEN}
\end{axis}
\begin{axis}[at={(5.3cm,0)}, width=5.4cm, height=4.6cm, title={Whispered audio\\called speech (\%)}, title style={font=\footnotesize, align=center},
  xtick={0.2,0.3,0.4,0.5}, xlabel={Threshold}, xlabel style={font=\footnotesize}, ticklabel style={font=\scriptsize}, ymin=0, ymax=100,
  axis x line*=bottom, axis y line*=left, ymajorgrids, major grid style={black!12}]
\addplot[figblue, line width=1.4pt, mark=*, mark size=1.8pt] coordinates {(0.2,51.7) (0.3,46.5) (0.4,41.8) (0.5,37.3)};
\addplot[figorange, line width=1.4pt, mark=square*, mark size=1.8pt] coordinates {(0.2,99.8) (0.3,97) (0.4,88.6) (0.5,79)};
\end{axis}
\begin{axis}[at={(10.6cm,0)}, width=5.4cm, height=4.6cm, title={No-speech clips\\with text (of 10)}, title style={font=\footnotesize, align=center},
  xtick={0.2,0.3,0.4,0.5}, xlabel={Threshold}, xlabel style={font=\footnotesize}, ticklabel style={font=\scriptsize}, ymin=0, ymax=10,
  axis x line*=bottom, axis y line*=left, ymajorgrids, major grid style={black!12}]
\addplot[figblue, line width=1.4pt, mark=*, mark size=1.8pt] coordinates {(0.2,0) (0.3,0) (0.4,0) (0.5,0)};
\addplot[figorange, line width=1.4pt, mark=square*, mark size=1.8pt] coordinates {(0.2,2) (0.3,2) (0.4,2) (0.5,2)};
\end{axis}
\end{tikzpicture}
\caption{Self-hosted Whisper with the VAD dropping audio, at four drop thresholds. Left: whispered WER (sending all of the audio gives 10.1\%). Middle: share of the whispered audio's 32\,ms windows the VAD calls speech. Right: no-speech clips on which any text was written. One run per threshold.}
\label{fig:vadsweep}
\end{figure}

\paragraph{The threshold sweep.} Both dropping setups were repeated at thresholds 0.2, 0.3 and 0.4 on the whispered, fan-on and no-speech clips (Figure~\ref{fig:vadsweep}). Silero's whispered WER falls only from 51.4\% to 41.1\%, because even at 0.2 it calls 52\% of whispered audio speech; TEN stays at 10.0--10.7\%. Fan-on WER moves by at most 2.1 points, and no-speech clips with text stay at 0 of 10 for Silero and 2 of 10 for TEN.

\end{document}